\documentclass[journal]{IEEEtran}

\usepackage[pdftex]{graphicx}
\graphicspath{{figure/pdf/}{figure/eps/}}
\usepackage[dvipsnames]{xcolor}

\usepackage[T1]{fontenc}
\usepackage{amsmath}
\usepackage{amsfonts}
\usepackage{bm}
\usepackage{algorithm}
\usepackage{algpseudocode}
\usepackage{multirow}
\usepackage{verbatim}

\usepackage{array}

\ifCLASSOPTIONcompsoc
	\usepackage[caption=false,font=normalsize,labelfont=sf,textfont=sf]{subfig}
\else
	\usepackage[caption=false,font=footnotesize]{subfig}
\fi

\newcommand{\X}{\mathbf{X}}
\newcommand{\Y}{\mathbf{Y}}
\newcommand{\Z}{\mathbf{Z}}
\newcommand{\M}{\mathbf{M}}
\newcommand{\W}{\mathbf{W}}
\newcommand{\K}{\mathbf{K}}
\newcommand{\bLambda}{\mathbf{\Lambda}}

\newcommand{\T}{\mathbf{T}}
\newcommand{\y}{\mathbf{y}}
\newcommand{\z}{\mathbf{z}}
\newcommand{\ba}{\mathbf{a}}

\newcommand{\V}{\mathbf{V}}
\newcommand{\I}{\mathbf{I}}
\newcommand{\G}{\mathbf{G}}

\newcolumntype{C}[1]{>{\centering\arraybackslash}m{#1}}

\begin{document}
\title{Accurate and Scalable Image Clustering Based On Sparse Representation of Camera Fingerprint}
\author{Quoc-Tin~Phan, Giulia~Boato, Francesco~G.~B.~De~Natale%
	\thanks{The authors are with the Department of Information Engineering and Computer Science, University of Trento, Trento 38123, (email: \{quoctin.phan, giulia.boato, francesco.denatale\}@unitn.it) and also affiliated to the National Inter-University Consortium for Telecommunications (CNIT).}
}

\maketitle

\IEEEtitleabstractindextext{
	\begin{abstract}
		Clustering images according to their acquisition devices is a well-known problem in multimedia forensics, which is typically faced by means of camera Sensor Pattern Noise (SPN). Such an issue is challenging since SPN is a noise-like signal, hard to be estimated and easy to be attenuated or destroyed by many factors. Moreover, the high dimensionality of SPN hinders large-scale applications. Existing approaches are typically based on the correlation among SPNs in the pixel domain, which might not be able to capture intrinsic data structure in union of vector subspaces. In this paper, we propose an accurate clustering framework, which exploits linear dependencies among SPNs in their intrinsic vector subspaces. Such dependencies are encoded under sparse representations which are obtained by solving a LASSO problem with non-negativity constraint. The proposed framework is highly accurate in number of clusters estimation and image association. Moreover, our framework is scalable to the number of images and robust against double JPEG compression as well as the presence of outliers, owning big potential for real-world applications. Experimental results on Dresden and Vision database show that our proposed framework can adapt well to both medium-scale and large-scale contexts, and outperforms state-of-the-art methods.
	\end{abstract}
	
	}

\IEEEdisplaynontitleabstractindextext
%
% INTRODUCTION
%
\section{Introduction}
\label{sec:INTRODUCTION}

Getting information about the camera used to acquire an image provides forensic analysts with important cues to counterfeit digital crime. Occasionally, such information can be extracted from the attached metadata, e.g., the Exif header. However, this may be unavailable or can be easily modified or swept out even by non-experts. A more interesting option would be to detect the source directly from the image data. To this purpose, it has been observed that digital images intrinsically contain a Sensor Pattern Noise (SPN) caused by sensor imperfections of the capturing device \cite{Lukas2006,Chen2008}. Just like human fingerprints, SPN uniquely identifies the acquisition source and can be considered as a \textit{camera fingerprint}. In blind scenarios, where only a set of unsourced images are given, such fingerprint can reveal images that share the same source. We refer to this task as image clustering by source camera. Further investigations, for instance detecting how many cameras a suspect owns, or how likely an image is taken from the suspect's camera, can derive from clustering results.  Indexing images by source camera also leads to direct applications in large-scale image retrieval.

In order to properly estimate the camera fingerprint, a number of smooth and uniformly bright images should be collected \cite{Chen2008}. Unfortunately, this requirement is usually not fulfilled in a blind scenario, where all images are unlabeled and no assumption can be made about the visual content. Consequently, SPN can just be coarsely approximated from the noise residual of a single image, which contains not only the pattern noise but also various other noise sources, such as shot noise and noise resulting from lossy compression or other filtering. Different methods have been proposed to enhance the SPN estimation and matching, such as averaging \cite{Lukas2006}, PCA+LDA \cite{Li2015}, spectrum equalization \cite{Lin2016_2}. All those methods, however, require a labeled training set, making them unsuitable for image clustering by source camera in unsupervised scenarios.

Existing unsupervised techniques are typically based on the normalized correlation among SPNs, used as a similarity measure, whose degree of reliability is limited by the impact of multiple noise sources. In \cite{Bloy2008}, an image is assigned to a group if the correlation between its noise residual and the relevant centroid exceeds a threshold, approximated by a quadratic model. Markov Random Fields are applied in \cite{Li2010_2,Li2017} to iteratively assign a class label to an image based on the consensus of a small set of SPNs, called membership committee. This raises another problem on how to choose a good committee, especially on asymmetric datasets where cluster cardinalities are unbalanced. In \cite{Caldelli2010, GarciaVillalba2015, Fahmy2015}, a hierarchical partition - a binary tree containing singleton clusters as leaf nodes and whose root node is a cluster containing all data points - is built by hierarchical clustering. The major problem of existing hierarchical approaches is the sensitivity to noise and outliers, as a wrong assignment might result in the propagation of errors to higher tiers. Multiclass spectral clustering is applied in \cite{Liu2010} to partition an undirected graph of unsourced images. The algorithm starts with two clusters and stops when it finds a cluster containing only one member. This stopping condition is heuristic, and as been improved by using normalized cut in \cite{Amerini2014}. Recently, in \cite{Marra2016,Marra2017}, multiple base partitions are obtained on top of multiple binarized undirected graphs and then combined to form a complete clustering solution.

Another important problem that has to be taken into account is scalability. In practical applications, often the clustering has to be applied to large databases, containing huge numbers of high-resolution images. To the best of our knowledge, only the method in \cite{Lin2017} addresses large-scale clustering of camera fingerprints, where the main idea is to split the dataset into small batches, which can be efficiently loaded on RAM, and to apply a coarse-to-fine clustering.

In the present work, we propose a clustering framework that exploits linear dependencies among SPNs in their intrinsic vector subspaces. Such dependencies are encoded under sparse representations, which are solutions of a constrained \textproc{LASSO} problem. Well-known clustering methods can then be applied on top of sparse representation matrix to obtain the final segmentation. Our framework is scalable despite of the complexity of \textproc{LASSO} thanks to a divide-and-conquer mechanism, which allows clustering on large datasets. Experimental tests on medium-scale and large-scale contexts exhibit advantages of sparse representations on clustering performance. The robustness of our proposed framework is demonstrated against the presence of outliers and double JPEG compression.
A similar approach has been preliminarily described in \cite{Phan2017}. Differently from \cite{Phan2017}, here we impose a non-negativity constraint that provides the \textit{interpretability} of solutions, thus allowing the extension to large-scale contexts. 

The rest of the paper is organized as follows: in Section \ref{sec:PRELIMINARIES} we describe the extraction of SPNs and the sparse subspace clustering method; in Section \ref{sec:PROPOSED_METHOD} we present the proposed optimization problem and its solution, as well as a clustering framework for large-scale datasets. Finally, discussions on computational complexity and extensive experimental analysis are provided in Section \ref{sec:COMPLEXITY_ANALYSIS} and \ref{sec:EXPERIMENTS}, respectively.

%
% RELATED WORK
%
%\section{Related Work}\label{sec:RELATED_WORK}
%
\section{Preliminaries}
\label{sec:PRELIMINARIES}

Throughout the paper, elements of a matrix or a vector are indicated by subscripts and single subscript $\M_i$ denotes the $i^\text{th}$ column of $\M$. With $\text{diag}(\M)$ we denote the matrix containing only the diagonal of $\M$, all other entries being set to zero. Infinity norm of $\y$ is defined as $\|\y\|_\infty = \underset{i}{\max}\, |\y_i|$, while $\ell_1$ and $\ell_2$ norms of vector $\y$ are denoted as $\|\y\|_1$ and $\|\y\|_2$, respectively. For matrices, $\ell_1$ norm, Frobenius norm and infinity norm are respectively defined as: $\|\M\|_1 = \underset{i,j}{\sum}\, |\M_{ij}|$, $\|\M\|_F = \sqrt{\underset{i,j}{\sum} \, \M_{ij}^2}$, $\|\M\|_\infty = \underset{i,j}{\max} \, |\M_{ij}|$.

\subsection{Sensor Pattern Noise}
\label{sec:SPN}  
Given a grayscale image $\Y$, its noise residual $\W$ can be extracted by a denoising filter.
A simplified model of $\W$ can be expressed as follows \cite{Chen2007,Chen2008}: 
\begin{IEEEeqnarray}{rCl}
	\W &=&  \T\Y\K + \mathbf{\Xi} \text{,}
	\label{W_estimate}
\end{IEEEeqnarray}
where $\mathbf{\Xi}$ is a the matrix of independently and identically distributed (i.i.d) Gaussian random variables, $\T$ is an attenuation matrix, and $\K$ is referred to as Photo-Response Non-Uniformity (PRNU).
%
% Sparse Representation and Normalized Correlation
%
In theory, PRNU can be used to cluster images with respect to the acquisition device. However, in a blind scenario this is difficult due to two main problems. First, the Cramer-Rao Lower Bound on the variance of PRNU estimate indicates that a number of smooth and bright (but not saturated) images are required for each camera \cite{Chen2008} and this condition is hardly satisfied. 
Indeed, the only available information is the noise residual $\W$ for each image, which contains not only the PRNU but also the additive noise $\mathbf{\Xi}$, which limits the reliability of traditional similarity measures used in conventional clustering algorithms. Several methods have been proposed for SPN enhancement \cite{Li2010, Lin2016_2} but it has been confirmed by \cite{Lin2017} that such methods are not suitable for unsupervised setting.
Second, the dimension of camera fingerprints is usually high, due to the high resolution of camera sensors, thus their clustering requires huge computation and memory as long as the number of data increases.  

Existing approaches use normalized correlation to measure the similarity between two flattened fingerprints $\ba,\mathbf{b}$ of dimension $d$:
\begin{IEEEeqnarray}{rCl}
	\rho(\ba,\mathbf{b}) &=& \frac{\sum_{i=1}^{d} \left( \ba_i - \bar{\ba} \right)\left( \mathbf{b}_i - \bar{\mathbf{b}} \right)}{\sqrt{\sum_{i=1}^{d}\left( \ba_i - \bar{\ba} \right)^2} \sqrt{\sum_{i=1}^{d}\left( \mathbf{b}_i - \bar{\mathbf{b}} \right)^2}}\text{,}
	\label{corr_definition}
\end{IEEEeqnarray}
where scalars $\bar{\ba}, \bar{\mathbf{b}}$ are the mean values of $\ba$ and $\mathbf{b}$, respectively. Without loss of generality, if $\ba, \mathbf{b}$ are normalized to have zero mean and unit norm, Eq. (\ref{corr_definition}) simply becomes:
\begin{IEEEeqnarray}{rCl}
	\rho(\ba,\mathbf{b}) &=& \sum_{i=1}^{d} \ba_i \mathbf{b}_i \IEEEnonumber \text{,}
\end{IEEEeqnarray}
which represents the cosine similarity between $\ba$ and $\mathbf{b}$. 

\subsection{Sparse Representation}

Given a set of data points arranged into the columns of a matrix $\X  \in \mathbb{R}^{d \times n}$, a data point $\mathbf{y}$ can be expressed as a linear combination of the columns of $\X$. A \emph{sparse} combination reveals columns in the same subspace which $\y$ happens to lie into. Sparse Subspace Clustering (SSC) \cite{Elhamifar2013} finds a sparse representation of $\mathbf{y}$ by solving the following optimization problem:
\begin{equation}
	\underset{\z}{\text{minimize}} \quad \|\z\|_1 \quad \text{subject to} \quad  \X\z = \y \text{.} 
	\label{eq:naive_ssc}
\end{equation}

If columns of $\mathbf{X}$ are contaminated by noise or not well distributed, $\X\z = \y$ might never be reached. Works in \cite{Soltanolkotabi2014,Wang2016} have shown that SSC can deal with noisy data if Eq. (\ref{eq:naive_ssc}) is reformulated as a LASSO problem:
\begin{equation}
	\underset{\z}{\text{minimize}} \quad \|\X\z - \y\|_2^2 + \gamma \|\z\|_1 \text{, }
	\label{eq:ssc}
\end{equation}
where $\gamma \geq 0$ is a regularization hyperparameter. 

\begin{figure}
	\centering
	\includegraphics[width=.45\textwidth]{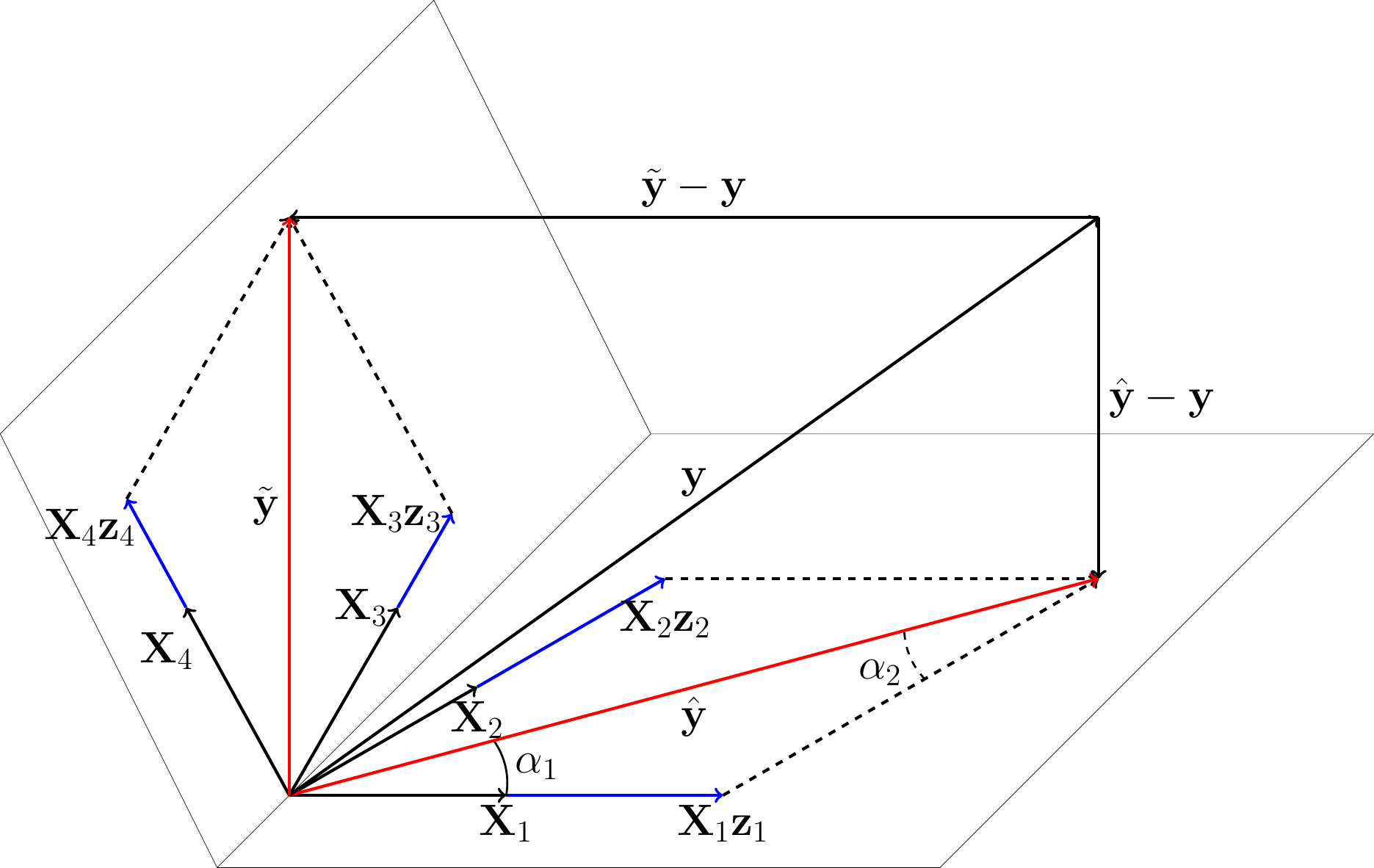}
	\caption{Geometric interpretation of solution of SSC. }
	\label{fig:geometric_ssc}
\end{figure}

Let us show a simple example to interpret the solutions of SSC. As illustrated in Figure \ref{fig:geometric_ssc}, we have $\X = \left[ \X_1, \X_2, \X_3, \X_4 \right] \in \mathbb{R}^{3\times4}$. We assume there are two subspaces spanned by $\left[\X_1, \mathbf{X}_2\right]$ and $\left[\mathbf{X}_3, \X_4\right]$. Without the regularization term, $\y$ can be expressed as linear combination of any $3$ columns of $\X$. However, we would like to assign $\y$ to the closest subspace spanned by $\left[\X_1, \X_2\right]$, i.e., $\| \hat{\y} - \y \|_2 < \| \tilde{\y} - \y \|_2$, a preferable solution should satisfy:
\begin{equation}
	\hat{\y} = \z_1\X_1 + \z_2\X_2 \IEEEnonumber \text{.}
\end{equation}
Such solution can be reached if we encourage the sparseness of $\z$ by penalizing $\|\z\|_0$. Unfortunately, $\ell_0$ optimization is usually intractable due to its non-convex and combinatorial nature. $\|\z\|_1$ can be used instead, being a good approximation of $\|\z\|_0$ \cite{Donoho2004}.
Exploiting the $\ell_1$ regularization term as in Eq. (\ref{eq:ssc}) and properly selecting $\gamma$, SSC finds a sparse solution that minimizes the reconstruction error.

	\subsection{Motivation}
	Even if two fingerprints come from the same camera, their normalized correlation is very weak due to the presence of irrelevant pixels. Only a subset of pixels are supposed to be relevant for each camera. Finding the subset of relevant pixels is connected to finding the subspace where fingerprints of a camera happen to lie in. Noticeable efforts in the direction of dimensionality reduction, such as fingerprint digest \cite{Goljan2010} or random projection \cite{Valsesia2015}, cannot be applied in clustering problems. In fact, \cite{Valsesia2015} considers a single subspace, while different cameras have different subsets of relevant pixels. On the other hand, fingerprint digest is composed only by saturated values of the reference fingerprint, which can be extracted only if the common source of images is known.
	
	SSC has been exploited to find structure of data in their intrinsic subspaces. 
	Indeed, SSC formulated under \textproc{LASSO} problem works in broad conditions: theoretical guarantees had been provided when subspaces intersect \cite{Soltanolkotabi2011}, or in the presence of additive noise \cite{Soltanolkotabi2014} even if the level of noise is higher than level of signal  \cite{Wang2016}, demonstrating that in practice SSC can reliably recover cluster memberships. Moreover, SPNs are known as compressible signals: low-dimensional representations of SPNs are found in \cite{Li2015,Rao2017,LI2018,Valsesia2015, Valsesia2015_2}, thus implying the existence of subspaces that can well represent them.
	
	Finally, we observe that the residual $\mathbf{W}$ is a noisy estimate of true camera fingerprint, thus the distributions of intra-class and inter-class correlations computed on $\mathbf{W}$ are heavily overlappped, making clustering algorithms less accurate. This challenge raises the need to eliminate inter-class data relationships and obtain unambiguous underlying data structure. By leveraging sparsity, SSC expresses each data point by a few linear relationships with other data points and extracts unambigous representation, which is essential for this problem.
%
% Proposed Method
%
\section{Proposed Method}
\label{sec:PROPOSED_METHOD}
In this section, we present our clustering framework including three steps:
\begin{itemize}
	\item \textit{Fingerprint extraction and normalization} (Section \ref{sec:FINGERPRINT_EXT_NORM}): given a set of color images as input, we extract, refine and normalize the corresponding noise residuals.
	\item \textit{Proposed optimization} (Section \ref{sec:OPT_SOLVING}): we present a constrained optimization problem to retrieve sparse and interpretable solutions.
	\item  \textit{Extension to large-scale contexts} (Section \ref{sec:LARGE-SCALE CLUSTERING}): we design a divide-and-conquer mechanism enabling large-scale clustering.
\end{itemize}
%
% Fingerprint Extraction and Normalization
%
\subsection{Fingerprint Extraction and Normalization}
\label{sec:FINGERPRINT_EXT_NORM}
In this study, we do not make any assumption on image content, but we simply filter out dark images (if any) since dark or textured images are inappropriate for fingerprint estimation. An image is considered dark if more than $75\%$ of pixels have values in $[0,80]$.

A noise residual $\W^c$, $c \in \{\text{red, green, blue\}}$ is extracted from each color channel of $\Y$, by exploiting the wavelet-based denoising filter used in \cite{Lukas2006,Chen2007,Chen2008,Goljan2010}, and then converted to one-channel noise residual. To further suppress non-unique artifacts caused by color interpolation of demosaicing algorithms and standard JPEG compression, we subtract from each row the mean of rows and from each column the mean of columns, and transform the obtained noise residual $\W$ into a one-dimensional unit-norm signal. We then obtain a data matrix $\X \in \mathbb{R}^{d \times n}$, $n$ being the number of fingerprints, and $d$ the number of pixels.
%
% Optimization Solving
%
\subsection{Proposed optimization}
\label{sec:OPT_SOLVING}

SSC learns a sparse representation $\z$ of $\y$, whose non-zero entries indicate data points closest to the orthogonal projection of $\y$ onto the relevant subspace. We can \emph{interpret} the magnitude of $\z_i$ as a similarity measure: the closer $\X_i$ is to $\hat{\y}$, the more it contributes to the reconstruction of $\y$, resulting in a larger value of $\z_i$. Back to the example in Figure \ref{fig:geometric_ssc}, denoted as $\alpha_i=\angle \left(\X_i, \hat{\y}\right)$ the angle between $\hat{\y}$ and $\X_i$, it is easily to see that if $\alpha_i < \alpha_j$ then $|\z_i| > |\z_j|$:
\begin{equation}
	\alpha_i < \alpha_j 
	\Leftrightarrow	 \text{cos}(\alpha_i) >\text{cos}(\alpha_j) 
	\Leftrightarrow	 \|\z_i\X_i \|_2  > \|\z_j\X_j \|_2 \IEEEnonumber \\
\end{equation}
\begin{equation}
	\Leftrightarrow 	  \left |\z_i \right| > \left| \z_j \right| \text{(since $\|\X_i\|_2 = 1$).} \IEEEnonumber
\end{equation}

\begin{figure}
	\centering
	\includegraphics[width=0.8\linewidth]{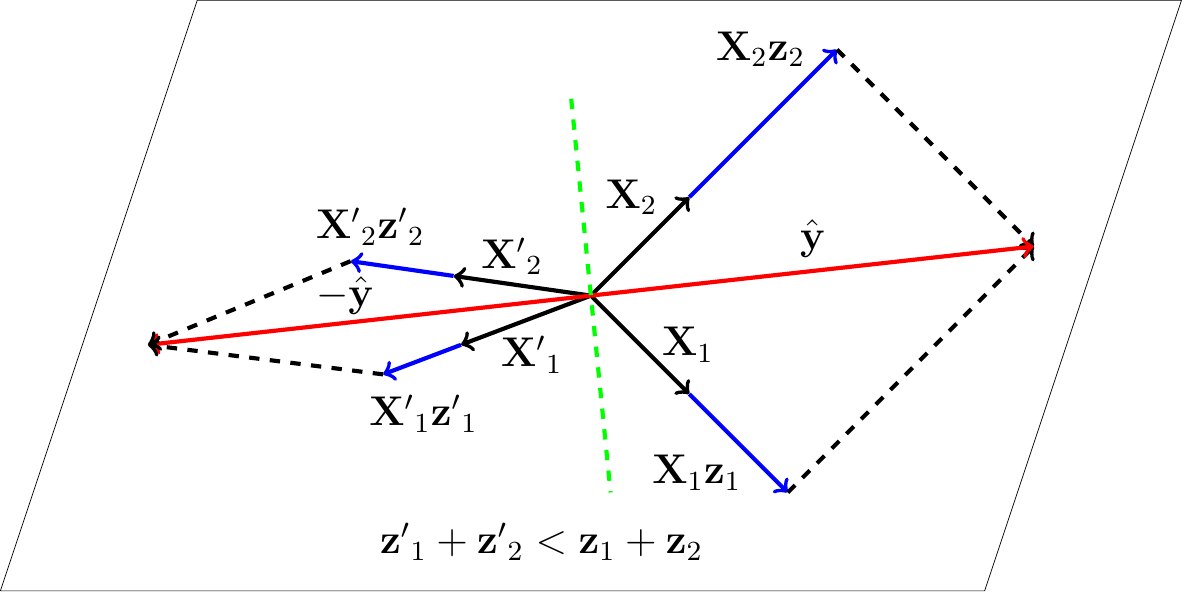}
	\caption{Geometric interpretation of negative solution of SSC.}
	\label{fig:geometric_ssc_counter}
\end{figure}

Thus, $\left | \z_i \right |$ is inversely propotional to $\alpha_i$. The $\ell_1$ regulari\-zation term encourages the sparseness of $\z$, whose non-zero entries should indicate data points \emph{closest} to $\hat{\y}$. Due to the nature of $\ell_1$ norm, however, negative and positive contributions are weighted equally. We provide in Figure \ref{fig:geometric_ssc_counter} an example where solution $\tilde{\z}=[-\z'_1, -\z'_2]^T$ might be chosen instead of $\z=\left[\z_1, \z_2\right]^T$ as it is possible that $\left \|\tilde{\z}\right\|_1 < \left \|\z\right\|_1$. Note, in this example $\z_i > 0$ and $\z'_i > 0$, thus $\tilde{\z}$ contains negative entries. This is an unexpected solution since $\X'_1, \X'_2$ lie in another half-space, i.e., $\angle(\X'_1, \hat{\y}) > \pi/2$ and $\angle(\X'_2, \hat{\y}) > \pi/2$. In order to avoid this solution, all entries $\z_i$ can be constrained to non-negative values. Therefore, the optimal solution reveals data points lying in the subspace closest to $\y$ and correlated to the orthogonal projection of $\y$ on that subspace. Another intesting property is that if $\z_i \geq 0$, $\z_j \geq 0$ and $\alpha_i < \alpha_j$ then $\z_i > \z_j$. This motivates us to impose a non-negativity constraint on the optimization problem.

For each column $\X_i$, we expect to learn a sparse representation $\Z_i$ such that $\X_i = \X\Z_i$. To obtain a meaningful representation, a column should not be expressed by itself, thus requiring the constraint ${\Z_{ii} = 0}$. Accordingly, we have to solve the following optimization problem:
\begin{IEEEeqnarray}{rl} \label{proposed_opt}
	\underset{\Z}{\text{minimize}} &\quad \frac{1}{2}\left \| \X\Z-\X \right \|_F^2 + \gamma \left \|\Z \right\|_1 \IEEEnonumber \\
	\text{subject to} &\quad \text{diag}(\Z) = 0, \Z \geq 0 \text{,}
\end{IEEEeqnarray}
where $\gamma > 0$ is the regularization hyperparameter.

Many research efforts have been spent in solving the unconstrained version of Eq. (\ref{proposed_opt}) \cite{Yang2013}. The $\ell_1$ minimization problem does not have an analytical solution; its solution instead has to be obtained numerically. Among the proposed algorithms, Augmented Lagrange Multiplier (ALM) generally converges faster under a wide range of data \cite{Yang2013}. In this paper, we adopt Alternating Direction Method of Multipliers (ADMM) \cite{Boyd2011} to solve the problem in Eq. (\ref{proposed_opt}), which couples the fast convergence of ALM with the decomposibility property, which is fundamental for distributed implementation in large-scale problems. 
ADMM introduces a complementary variable $\V$ and re-formulates the unconstrained version of Eq. (\ref{proposed_opt}) into the following equivalent form:
\begin{IEEEeqnarray}{rl} \label{eq:lasso_admm}
	\underset{\Z,\V}{\text{minimize}} &\quad  \gamma\| \V \|_1 +  \frac{1}{2}\| \X\Z-\X\|^2_F \IEEEnonumber \\
	\text{subject to} &\quad \Z=\V \text{.}
\end{IEEEeqnarray}

Here, decomposibility means that $\Z$ and $\V$ can be updated separately, possibly on a distributed system, thus constraints in Eq. (\ref{proposed_opt}) can be imposed on $\V$. They are enforced during $\V$ update by Euclidean projections which are much simpler than ALM.
The augmented Lagrangian form of Eq. (\ref{eq:lasso_admm}) is 
\begin{IEEEeqnarray}{rl}
	\mathcal{L}_\eta (\Z,\V,\bLambda) {}={}& \gamma \|\V\|_1 + \frac{1}{2}\| \X\Z-\X\|^2_F + \langle \bLambda, \Z-\V \rangle \IEEEnonumber\\
	&+\frac{\eta}{2} \|\Z-\V\|_F^2 \text{.}\IEEEnonumber 
\end{IEEEeqnarray}
where $\bLambda \in \mathbb{R}^{n \times n}$ is the Lagrangian multiplier and $\eta>0$ is the augmented Lagrangian hyperparameter. ADMM iteratively optimizes $\mathbf{Z,V}$ in an alternate fashion, by keeping one variable fixed and updating the others:
\begin{IEEEeqnarray}{rCl}
	\Z^{t+1} &=& \arg \underset{\Z}{\min}\; \mathcal{L}_\eta \left(\Z, \V^t, \bLambda^t \right) \text{,} \label{eq:z_update} \IEEEnonumber\\
	\V^{t+1} &=& \arg \underset{\V}{\min}\; \mathcal{L}_\eta \left(\Z^{t+1},\V,\bLambda^t \right) \text{,} \IEEEnonumber \\
	\bLambda^{t+1} &=& \bLambda^t + \eta\left(\Z^{t+1}-\V^{t+1}\right) \text{.}\IEEEnonumber
\end{IEEEeqnarray}

It is straightforward to demonstrate that $\Z$ can be  updated by solving the linear equation:
\begin{equation}
	(\X^T\X + \eta\I)\Z = (\X^T\X - \bLambda + \eta\V) \IEEEnonumber \text{,}
\end{equation}
using Cholesky decomposition of $\X^T\X + \eta\I$. On the other hand, solution of $\V$ at each iteration is obtained through soft thresholding operator $S$ defined as:
\begin{IEEEeqnarray}{rCl}
	S_\nu \left( a \right) &=& 
	\begin{cases}
		a - \nu \quad & a > \nu  \IEEEnonumber              \\
		a + \nu \quad & a < -\nu  \; \text{.} \IEEEnonumber \\
		0             & |a| \leq \nu                        
	\end{cases}
\end{IEEEeqnarray}

Details of its update are provided in Appendix \ref{app:v_update}. After $\V$ update, the two following operators are applied to project $\V$ into the feasible set of solutions:
\begin{IEEEeqnarray}{rlll}
	\Pi_{D}(\M_{ij}) {}={}&
	\begin{cases}
		\M_{ij} & i \neq j       \\
		0       & i = j \text{,} 
	\end{cases} \label{Pi_D} \\
	\Pi_{N}(\M_{ij}) {}={}&
	\begin{cases}
		\M_{ij} & \M_{ij} \geq 0       \\
		0       & \M_{ij} < 0 \text{.} 
	\end{cases} \label{Pi_N} 
\end{IEEEeqnarray}
The optimization procedure is reported in Algorithm \ref{alg:lasso_solving_admm}: it converges efficiently to an acceptable solution as  $\|\Z -\V \|_{\infty} \rightarrow 0$.

\begin{algorithm}
	\caption {Constrained LASSO}
	\label{alg:lasso_solving_admm}
	\begin{algorithmic}[0]
		\Procedure{Constrained\_Lasso}{$\mathbf{X}, \gamma, \eta$}
		\State \textbf{initialize}: $\mathbf{Z} \gets 0, \mathbf{V} \gets 0, \bLambda \gets 0, \varepsilon \gets 10^{-4}$
		\While {convergence condition is not satisfied} 
		\State Fix the others, update $\mathbf{Z}$
		\begin{flalign}
			\hspace{3.5em}&\mathbf{Z} {} \gets {} (\mathbf{X}^T\mathbf{X}+\eta \mathbf{I})^{-1}(\mathbf{X}^T\mathbf{X} - \bLambda + \eta\mathbf{V})& \IEEEnonumber
		\end{flalign}
		\State Fix the others, update $\mathbf{V}$
		\begin{flalign}
			\hspace{3.5em}&\V_{ij} {} \gets {} S_{\frac{\gamma}{\eta}}\left( \Z_{ij} + \frac{\bLambda_{ij}}{\eta} \right) &\IEEEnonumber
		\end{flalign}
		\vspace{-1.5em}
		\begin{flalign}
			\hspace{3.5em}&\V_{ij} {} \gets {} \Pi_{D}(\Pi_{N}(\V_{ij})) &\IEEEnonumber
		\end{flalign}
		\State Fix the others, update $\bLambda$:
		$\bLambda \gets \bLambda + \eta\left(\Z-\V\right)$
		\State Check convergence condition: $\|\Z - \V\|_{\infty} < \varepsilon$
		\EndWhile
		\State \Return $\mathbf{Z}$
		\EndProcedure
	\end{algorithmic}
\end{algorithm}
%
% Clustering
%

\begin{figure}[t]
	\centering
	\includegraphics[width=\linewidth]{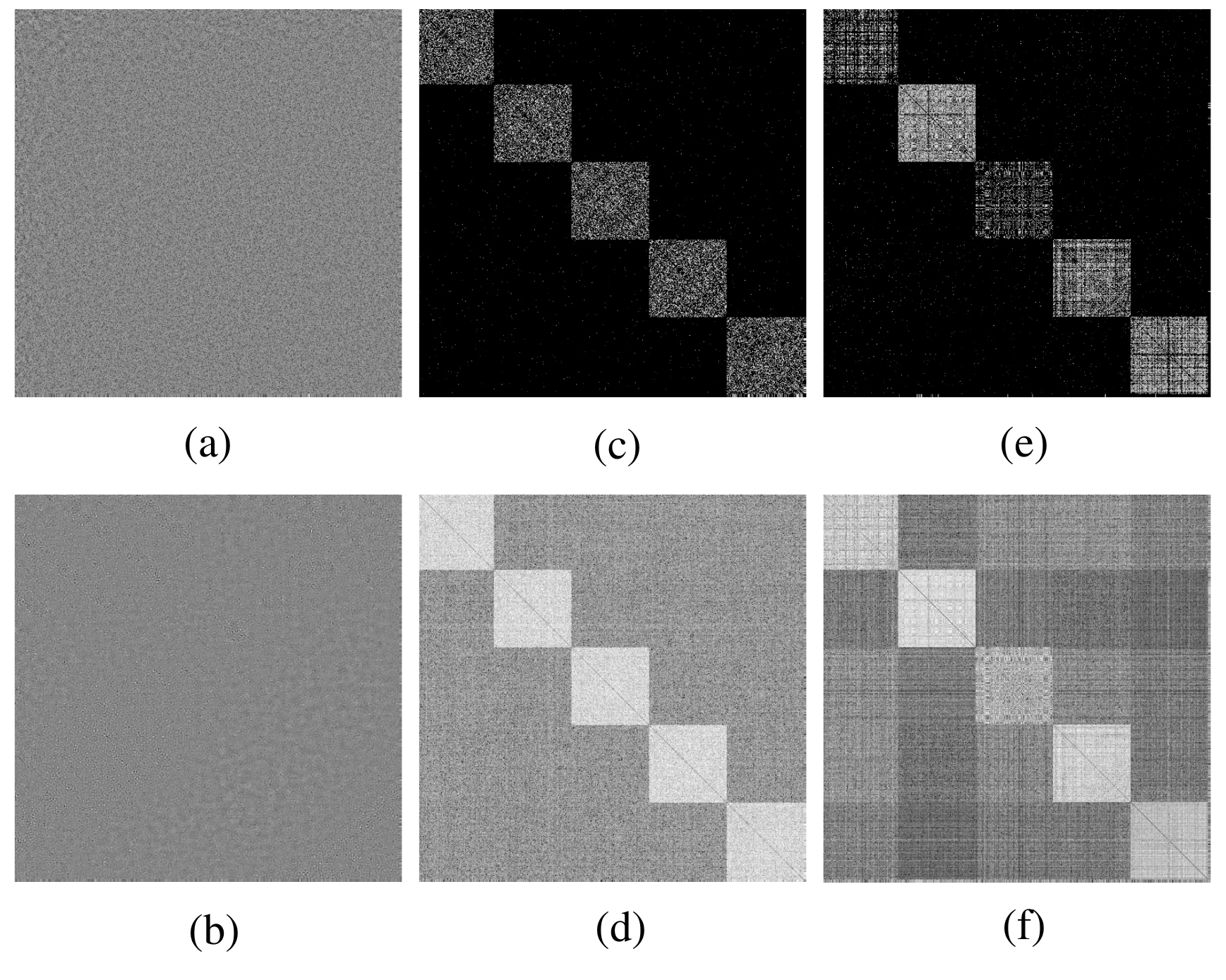}
	\caption{Visual comparison of sparse representation and dense representation (obtained by normalized correlation): (a) synthetic noise sample, (b) realistic noise sample, (c) sparse representation matrix of synthetic noise, (d) dense representation matrix of synthetic noise, (e) sparse representation matrix of realistic noise, (f) dense representation matrix of realistic noise.}
	\label{fig:sparse-vs-dense}
\end{figure}

To visually compare sparse representation and normalized correlation matrix, we conduct an analysis on synthetic noise and another one on realistic noise. Synthetic noise is extracted from images generated by the simple imaging model described in \cite{Chen2008} for smooth images (without the attenuation factor $\T$), that is $\Y = \Y^{(0)} + \Y^{(0)}\K + \mathbf{\Theta}$. The clean image $\Y^{(0)}$ is uniform, having pixel value of $0.9$ (relatively bright). $\K_{ij}$ and $\mathbf{\Theta}_{ij}$ are reasonably assumed as white Gaussian noise. As the signal $\K$ is generally weaker than $\mathbf{\Theta}$, the variance of $\K_{ij}$ is selected as $0.001$ and the variance of $\mathbf{\Theta}_{ij}$ is $0.1$ (for pixel values in $[0,1]$). We simulate the situation of $5$ cameras corresponding to $5$ different $\K$ patterns, considering $100$ images for each camera, thus resulting into $500$ different $\mathbf{\Theta}$ patterns. After that, we apply the same wavelet-based denoising filter to extract synthetic noise. A sample of extracted synthetic noise is depicted in Figure \ref{fig:sparse-vs-dense} (a). For the realistic setting, we select $5$ cameras from the Vision dataset \cite{Shullani2017}, $100$ images for each camera, and apply the same denoising procedure. In Figure \ref{fig:sparse-vs-dense} (b) an example of realistic noise is shown. We intentionally group noise residuals of the same camera so that the representation matrix is easily observable. We show sparse representation matrix of synthetic noise in Figure \ref{fig:sparse-vs-dense} (c), and of realistic noise in Figure \ref{fig:sparse-vs-dense} (e). The dense representation matrix in Figure \ref{fig:sparse-vs-dense} (d) and \ref{fig:sparse-vs-dense} (f) are obtained by computing pair-wise normalized correlation for synthetic noise and for realistic noise, respectively. Noticeably, solving the problem in Eq. (\ref{proposed_opt}) obtains meaningful representation where inter-class relations are effectively removed, revealing clearer block-diagonal structure compared to normalized correlation.

The sparse representation matrix captures asymmetric relationships among data points, i.e., $\Z_{ij} \neq \Z_{ji}$. For our clustering purpose, we build a weighted undirected graph $\G$ from $\Z$ as $\G=(\Z+\Z^T)/2$. To obtain the final segmentation, we apply the spectral clustering described in \cite{Ng2001} to partition $\G$ into $\kappa$ connected components or clusters. In practice, $\kappa$ can be inferred from the number of small eigenvalues. Therefore, we adopt an approach based on \emph{eigengap heuristic} \cite{vonLuxburg2007} to infer the number of clusters, similarly as \cite{Phan2017}.

In summary, the proposed algorithm learns sparse representations of each camera fingerprint. To avoid any confusion with ordinary SSC in \cite{Phan2017}, we denote Sparse Subspace Clustering with Non-negativity Constraint as SSC-NC. This approach provides a good instrument to discover structures on high-dimensional data, but shows a major drawback on scalability, since all data must be loaded in RAM. In Section \ref{sec:COMPLEXITY_ANALYSIS} the computational complexity of Algorithm \ref{alg:lasso_solving_admm} will be demonstrated to be in the order of $n^3$, $n$ being the number of fingerprints. Empirically, this is acceptable only for datasets with $n \leq 6000$.

%
% Large-Scale Clustering
%
\subsection{Large-scale sparse subspace clustering}
\label{sec:LARGE-SCALE CLUSTERING}
 
In this section we extend our methodology to cluster camera fingerprints in large-scale contexts, referring to such extension as large-scale SSC (LS-SSC). First, we address the memory issue using a \emph{divide-and-conquer} strategy, so that compact clusters could be discovered on small data batches. This process is followed by data re-cycling to increase the chance of discovering hidden clusters. Finally, we employ \emph{merging} and \emph{attraction} phases to finalize clustering results.
\begin{figure}[!t]
	\centering
	\includegraphics[width=\linewidth]{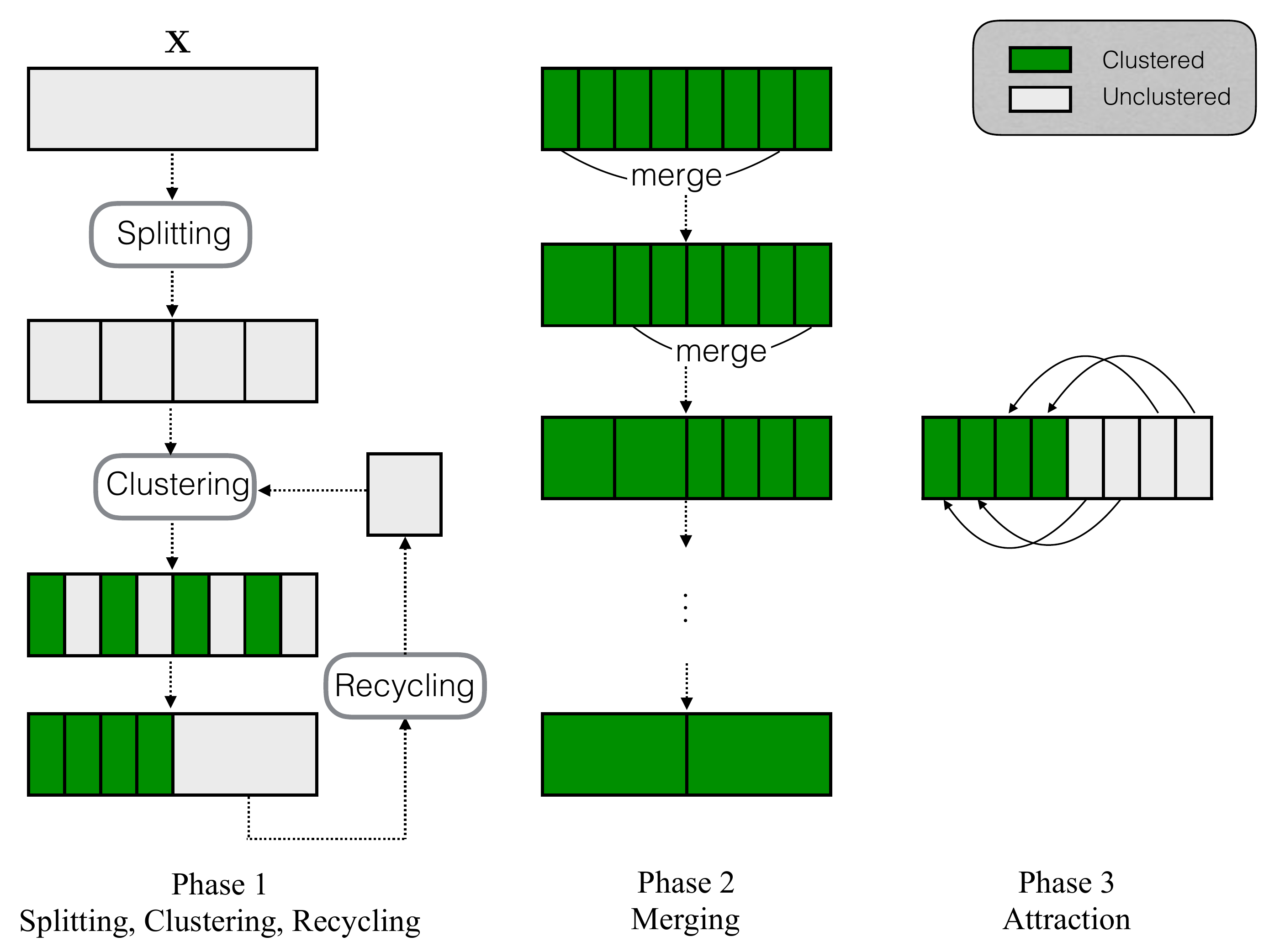}
	\caption{Schema of the proposed LS-SSC.}
	\label{fig:ls_schema}
\end{figure}
%
% Splitting, Clustering and Recycling
%
\subsubsection{Splitting, Clustering and Recycling}

The baseline strategy of our large-scale clustering is the divide-and-conquer paradigm, which breaks an intractable problem into several smaller tractable problems. We randomly split the set of all fingerprints $\mathcal{X}$ into $B$ batches of equal size, $\mathcal{X} = \left\{\mathcal{X}^l\right\}_{l=1,\ldots,B}$, where $B$ is originally set to $\lceil{\frac{n}{p}}\rceil$ and $p$ is the batch size. Only one data batch at a time is loaded on RAM. We then apply Algorithm \ref{alg:lasso_solving_admm} on the data batch to learn sparse representations among fingerprints.

We hereby refer to \emph{cluster purity} as the quality of a cluster. A pure cluster should contain only fingerprints of the same camera. The main purpose of this phase is to extract small-size but pure clusters that can be later merged to form larger clusters. As a result of splitting, a fingerprint might not be well reconstructed by only fingerprints from the same camera. To minimize the reconstruction error, the algorithm might select fingerprints from multiple cameras. Such representations are considered as \emph{outliers}, similarly to \cite{You2017}.

Let us now consider a sparse representation matrix as a directed graph: outliers have connections to both outliers and inliers, while inliers have connections to inliers only. If we perform a random walk on the graph, the probability of ending at inliers is therefore higher than ending at outliers. We apply the random walk algorithm described in \cite{You2017} with $1000$ steps to acquire the state probabilities, we model such probabilities as a normal distribution, and we keep $80\%$ of the distribution as inliers, thus classifying the rest as outliers.

To guarantee the purity of clusters, we avoid spectral clustering, but we attempt to localize dense regions using the interpretability property of our sparse representation. Accordingly, large values indicate \emph{closest} fingerprints. Since our target is to discover small-size but pure clusters, we can further simplify the graph by retaining only $K$ largest entries on each column of the sparse representation matrix, and setting other entries to zero. Each remaining fingerprint is located in a region of $K$ nearest neighbors, and fingerprints in the same cluster should have common neighbors, forming a dense region.

After that, we apply \textproc{DBSCAN} \cite{Ester1996} to discover dense regions. This classical clustering technique is computationally feasible for large-scale datasets and does not require the number of clusters to be known. Two parameters need to be indicated as input of \textproc{DBSCAN}: radius $\epsilon$ and minimum number of neighbors $MinPts$. The radius $\epsilon$ should be selected on the basis of $K$ largest values on each column of the sparse representation matrix, while $MinPts$ must be smaller or equal to $K$. If $\epsilon$ is too small, this results in many clusters. Conversely, very limited number of clusters are discovered, complicating the recycling process. On the other hand, if $MinPts > K$, there is no cluster discovered by \textproc{DBSCAN}. We empirically found that setting $MinPts = K$ and $\epsilon$ equal to the mean of non-zero entries, allows discovering pure clusters.  

After clustering, we obtain the set of inliers and outliers, where inliers are used for the merging phase and outliers are fed to the recycling process. The aim of recycling is to combine outliers from each batch and feed them back to the clustering process, thus increasing the chance to discover hidden clusters. For that reason, clustering and recycling can be seen as an iterative procedure, as outlined in Algorithm \ref{alg:Splitting_Clustering_Recycling}.

\begin{algorithm}
	\caption {Splitting, clustering, recycling}
	\label{alg:Splitting_Clustering_Recycling}
	\begin{algorithmic}[0]
		\Procedure{Splitting\_Clustering\_Recycling}{}
		\scriptsize
		\State \textbf{input}: $\mathcal{X}, p, R, K, \gamma, \eta$ \Comment{$\mathcal{X}$: dataset, $p$: batch size, $R$: number of recycling steps, $K$: number of nearest neighbors}
		\State \textbf{output}: $\mathcal{X}_{\text{in}}$, $\mathcal{X}_{\text{out}}$ \Comment{set of clustered and unclustered fingerprints}
		\State $\mathcal{X}_\text{in} \gets \emptyset$
		% Splitting
		\State $B \gets \lceil{\frac{n}{p}}\rceil$
		\State Split $\mathcal{X}$ into $\left\{ \mathcal{X}^l \right\}_{l = 1,\ldots,B}$
		\State $\mathcal{X}^l_\text{out} \gets \emptyset$, \Comment{$l = 1,\ldots,B$}
		\For{$l=1 \to B$}
		\State $\mathcal{\tilde{X}}^l_\text{in}$, $\mathcal{\tilde{X}}^l_\text{out} \gets $  \textproc{Partition}($\mathcal{X}^l$, $K, \gamma, \eta$)
		\State Append $\mathcal{\tilde{X}}^l_\text{in}$ to $\mathcal{X}_\text{in}$ and append $\mathcal{\tilde{X}}^l_\text{out}$ to $\mathcal{X}^l_\text{out}$
		\EndFor
					
		% Recycling
		\State $t \gets B$, $\tilde{B} \gets B$
		\Repeat
		\State $\mathcal{X}^t_\text{out} \gets \emptyset$,  $\mathcal{X}^t \gets \emptyset$
		\For{$l = 1 \to \tilde{B}$}
		\State Pop out randomly $s^l =  \frac{\left | \mathcal{X}_\text{out}^l \right | \times p} {\sum_{i=1}^{\tilde{B}} \left | \mathcal{X}^i_\text{out} \right |}$ fingerprints from $\mathcal{X}^l_\text{out}$
		\State Append $s^l$ fingerprints to $\mathcal{X}^t $
		\EndFor
		\State $\mathcal{\tilde{X}}^t_\text{in}$, $\mathcal{\tilde{X}}^t_\text{out} \gets $  \textproc{Partition}($\mathcal{X}^t$, $K, \gamma, \eta$)
		\State Append $\mathcal{\tilde{X}}^t_\text{in}$ to $\mathcal{X}_\text{in}$ and append $\mathcal{\tilde{X}}^t_\text{out}$ to $\mathcal{X}^t_\text{out}$
		\State $t \gets t + 1$, $\tilde{B} \gets \tilde{B} + 1$
		\Until{$t \geq B+R$}
		\State $\mathcal{X}_\text{out} \gets \left\{ \mathcal{X}^l_\text{out} \right\}_{l=1,\ldots,B+R}$
		\normalsize
		\EndProcedure

		\Procedure{Partition}{}
		\scriptsize
		\State \textbf{input}: $\mathcal{X}, K, \gamma, \eta$ \Comment{$\mathcal{X}$: dataset, $K$: number of nearest neighbors}
		\State \textbf{output}: $\mathcal{X}_{\text{in}}$, $\mathcal{X}_{\text{out}}$ \Comment{set of clustered and unclustered fingerprints}
		\State Load fingerprints in $\mathcal{X}$ to $\X$ \Comment{$\X$: matrix of fingerprints}
		% clustering
		\State $\Z \gets$ \textproc{Constrained\_Lasso($\X, \gamma, \eta$)}
		\State Remove outliers, obtain $\tilde{\Z}$. Append outliers to $\mathcal{X}_\text{out}$
		\State Keep only $K$ largest entries on each column of $\tilde{\Z}$, obtain $\tilde{\Z}_\text{KNN}$
		\State Apply DBSCAN to discover clusters
		\State Append inliers to $\mathcal{X}_\text{in}$
		\State Append outliers to $\mathcal{X}_\text{out}$
		\normalsize
		\EndProcedure
					
	\end{algorithmic}
\end{algorithm}

%
% Merging
%
\subsubsection{Merging}
In the first phase, by increasing \emph{K} we obtain larger clusters at the expense of a lower cluster purity. Conversely, we yield small-size pure clusters. The latter is preferable, as small-size clusters (subclusters) can be merged efficiently to form larger subclusters.

Let  $\W^A$ and $\W^B$ be two noisy fingerprints of dimension $d$, i.e., two singleton subclusters, reasonably assumed to follow a normal distribution since the denoising filter extracts stationary Gaussian noise in wavelet domain (see Appendix A of \cite{Lukas2006}). $\K^A$ and $\K^B$ are the noise-free fingerprints residing in $\W^A$, $\W^B$. The merging problem can be formulated as a classical hypothesis test:
\begin{IEEEeqnarray}{rCl}
	\text{H}_0&:& \K^A \neq \K^B \text{,} \IEEEnonumber\\
	\text{H}_1&:& \K^A = \K^B = \K. \IEEEnonumber
\end{IEEEeqnarray}

Under null hypothesis, $\rho \left( \W^A, \W^B \right) \sim \mathcal{N}(0,\frac{1}{d})$ according to the Central Limit Theorem (CLT). Two subsclusters can be merged if their normalized correlation exceeds $\frac{1}{\sqrt{d}} Q^{-1}(P_{FA})$, where $Q(t)$ is the probability that a standard normal variable is larger than $t$ and $P_{FA}$ is the expected false alarm rate \cite{Goljan2010}. More generally, if each cluster contains more than one fingerprint, $\W^A$ and $\W^B$ represent respectively the subcluster centroids. Under alternative hypothesis, the correlation between $\W^A$ and $\W^B$ increases if the cardinality of each subcluster increases, as random noise is effectively suppressed by averaging. Obviously, the merging phase will be more reliable if one knows not only the null distribution but also the alternative distribution.

To determine the alternative distribution, \cite{ROCFridrich,Lin2017} established a parametric model with some statistical assumptions and determined model parameters. For instance, the true fingerprints are assumed to be additive noise \cite{ROCFridrich,Lin2017}, and the WGN of a camera presents always the same variance \cite{ROCFridrich}.
Nevertheless, if those assumptions are not guaranteed, and usually they are not, parameter estimation becomes extremely difficult. 

We resort the merging problem into finding a threshold value $\tau$ that is able to exclude the null  hypothesis and to adapt to the variation of the alternative hypothesis, based on real data. This is achieved by taking into account the cardinality and intra-class correlation within each subcluster. Let $\X^A$ and $\X^B$ be the two matrices containing $n_A$ and $n_B$ fingerprints of each subcluster, and $\rho_A$ and $\rho_B$ be the intra-class correlation within these subclusters. We learn the threshold adaptiveness via linear regression:
\begin{IEEEeqnarray}{rCl}
	\mathcal{R} \left(n_A, n_B, \rho_A, \rho_B \right) &=& \left[n_A, n_B, \rho_A, \rho_B \right]\mathbf{w} + b  \IEEEnonumber \text{,}
\end{IEEEeqnarray}
where $\mathbf{w} \in \mathbb{R}^{4\times1}$ and $b \in \mathbb{R}$ are weights and bias, respectively. From real data, we calculate $\rho_A$, $\rho_B$ and the regression output $\mathcal{R}(\cdot)$ as follows: 
\begin{IEEEeqnarray}{rCl}
	\rho_A &=& \frac{1}{n_A(n_A-1)} \sum_{i=1}^{n_A} \sum_{j=1,j\neq i}^{n_A}\rho(\X^A_i, \X^A_j)  \IEEEnonumber \text{,}\\
	\rho_B &=& \frac{1}{n_B(n_B-1)} \sum_{i=1}^{n_B} \sum_{j=1,j\neq i}^{n_B}\rho(\X^B_i, \X^B_j)  \IEEEnonumber \text{,}\\
	\mathcal{R}(\cdot) &=&  \frac{\rho(\bar{\X}^A, \bar{\X}^B)}{2} \text{,}\IEEEnonumber
\end{IEEEeqnarray}
where $\bar{\X}^A, \bar{\X}^B$ are respectively two subcluster centroids, i.e., mean of columns in $\X^A$ and $\X^B$. The estimate of $\mathcal{R}(\cdot)$ is interpreted as the central value between mean of null and alternative distribution. The final regressor learnt from real data (we will mention this development set in Section \ref{sec:hyperparameter_selection}) has the form:
\begin{IEEEeqnarray}{rCl}
	\mathcal{R} \left(n_A, n_B, \rho_A, \rho_B \right) &=& 0.0016 \, n_A + 0.0016 \, n_B + \IEEEnonumber \\
	&  &2.2474 \, \rho_A + 2.2474 \, \rho_B - 0.0474 \text{.} \IEEEnonumber
\end{IEEEeqnarray}

Careful readers will notice that the regressor is symmetric in terms of cluster role, i.e., $\mathcal{R} \left(\right.n_A, n_B, \rho_A, \rho_B\left. \right)$\;=\;$\mathcal{R}\left(\right.n_B, n_A, \rho_B, \rho_A\left.\right)$. This is achieved by augmenting the training data with the role of two clusters exchanged. The threshold $\tau$ is finally calculated as:
\begin{IEEEeqnarray}{rCl}
	\tau &=& \max \left\{ \frac{1}{\sqrt{d}} Q^{-1}(P_{FA}), \mathcal{R} \left(\cdot \right) \right\}  \IEEEnonumber \text{,}
\end{IEEEeqnarray}
where $P_{FA}$ is chosen as $0.001$ ($0.1\%$ false alarm rate).

The merging phase is conducted as an iterative procedure, which respectively selects pairs of subclusters having maximum centroid correlation and compares them to $\tau$. If the correlation is larger than $\tau$, the two subclusters are merged and relevant information is updated. The algorithm stops when no more pairs of subclusters exist that satisfy the merging condition.

In practical cases, a good regressor might not be linear, but for $n_A, n_B$  within a reasonably small range, $\tau$ can be fitted by a linear function. Therefore, to calculate a reliable $\tau$, we set $n_A=\min\left\{\tilde{n}_A,50\right\}$ and $n_B = \min\left\{\tilde{n}_B, 50\right\}$ where $\tilde{n}_A,\tilde{n}_B$ are actual cluster cardinalities, and calculate $\rho_A$, $\rho_B$ using bounded sets of fingerprints. The quantity $50$ is suggested as a minimal cardinality for fingerprint estimation \cite{Lukas2005,Lukas2006}.

\begin{figure}[t!]
	\centering
	\includegraphics[width=\linewidth]{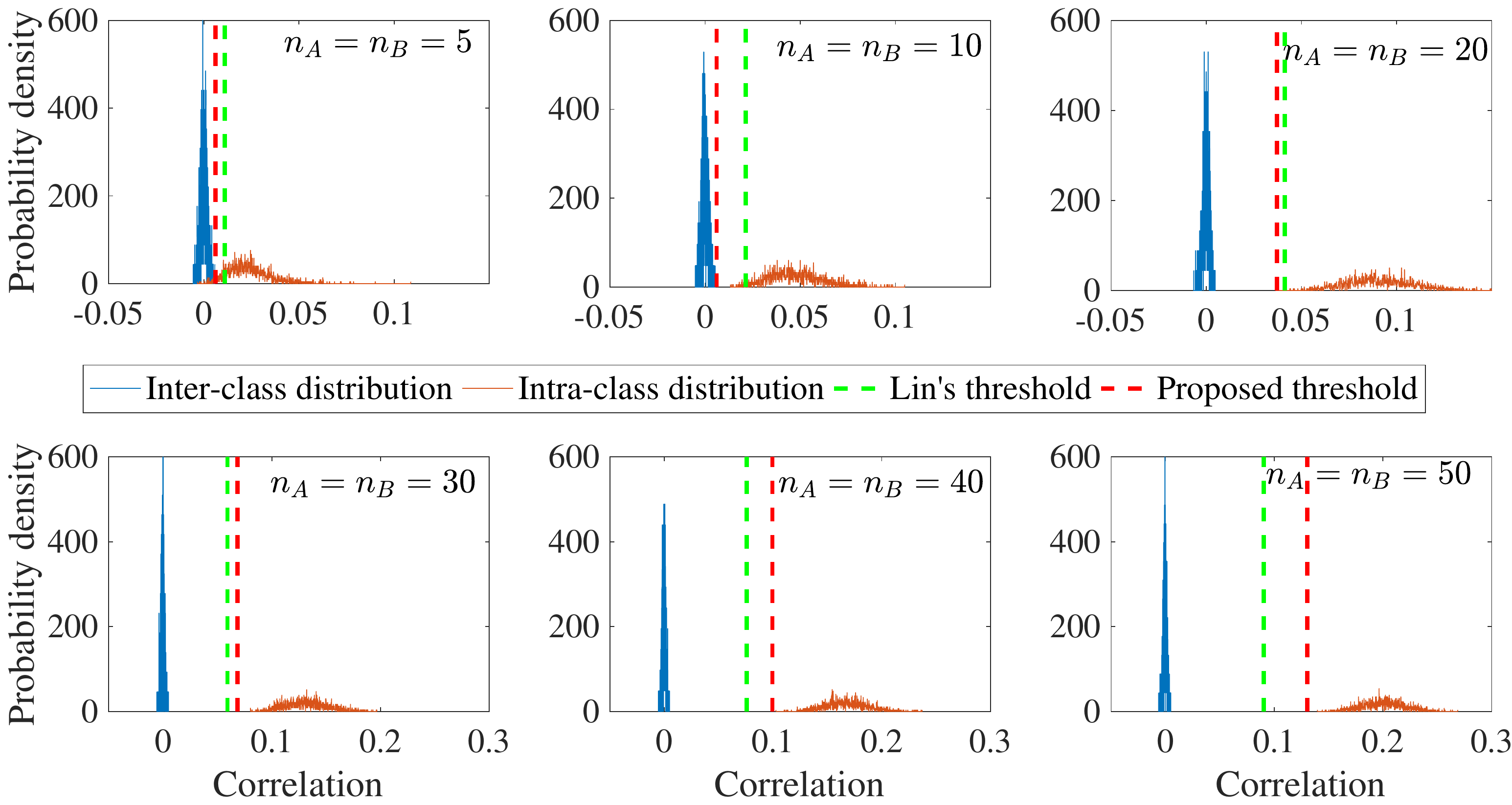}
	\caption{Comparison of the proposed threshold and Lin's threshold \cite{Lin2017} under two cameras: Kodak M1063 and Nikon CoolPix S710. Better viewed in color.}
	\label{fig:threshold_comparison_2}
\end{figure}

To demonstrate the effectiveness of the proposed threshold, we compare with Lin's threshold \cite{Lin2017}, which was previously shown to be superior than thresholds in \cite{Bloy2008} and \cite{Eklann2012}. We select two cameras, namely Kodak M1063 and Nikon CoolPix S710 from Dresden database \cite{Gloe2010}. We randomly split images of one camera into two parts to simulate two same-camera subclusters. Images from different cameras are used to create cross-camera subclusters. The process is replicated $2000$ times and intra-class and inter-class correlations are collected. Figure \ref{fig:threshold_comparison_2} shows how the proposed threshold and the threshold in \cite{Lin2017} (Lin's threshold) separate null and alternative distribution. When the cardinality of same-camera subclusters increases, the alternative distribution shifts towards the right, while the null distribution is centered at $0$. The proposed threshold consistently splits the two distributions, while Lin's threshold tends to be unnecessarily confident when two distributions are close. An interesting behavior of the proposed threshold and Lin's threshold is their adaptiveness to distribution shifting.
%
% Attraction
%
\subsubsection{Attraction}
In attraction phase, we assign remaining fingerprints to available clusters. Let us denote $\mathbf{C} = \left[ \bar{\mathbf{C}}_1, \bar{\mathbf{C}}_2, \ldots, \bar{\mathbf{C}}_L \right] \in \mathbb{R}^{d \times L}$ the matrix containing centroids of $L$ final clusters, and $\mathbf{X^\text{out}} \in \mathbb{R}^{d \times U}$ the data matrix containing $U$ unclustered fingerprints. Since the quality of camera fingerprints is generally non-homogeneous, cluster assignment should be performed for high-quality fingerprints first, in order to minimize assignment errors. The cluster membership $l, 1 \leq l \leq L$ of fingerprint $\mathbf{X}_i^\text{out}$, $1 \leq i \leq U$ is obtained iteratively by finding at each step the pair $l$ and $i$ such that $\rho(\X_i^\text{out}, \bar{\mathbf{C}}_l)$ is maximum  and greater than $Q^{-1}(P_{FA})$ which is the threshold used to exclude null hypothesis in merging phase. After being attracted $\X^\text{out}_i$ is discarded, otherwise it is labeled to as \emph{unclustered}.
Since the remaining fingerprints to be merged have been classified as outliers after recycling, we can expect that they are low-quality samples. Therefore, to reduce false alarm rate, the cluster centroid is updated only when its cardinality does not exceed $50$, consistently with the empirical value used in the merging phase. 

Eventually, we obtain the cluster memberships of camera fingerprints in a large-scale database  and a number of unclustered fingerprints. 

%
% Complexity Analysis
%
\section{Computational Complexity}
\label{sec:COMPLEXITY_ANALYSIS}
In this section we discuss on the time complexity of our proposed SSC-NC, LS-SSC and two recent works: correlation clustering with consensus (CCC) \cite{Marra2017} and Lin's large-scale method (Lin-LS) \cite{Lin2017}.

\textbf{SSC-NC}. SSC-NC is composed by \textproc{Constrained\_Lasso} and spectral clustering. \textproc{Constrained\_Lasso} consists of Cholesky decomposition, linear equation solving and soft thresholding. In the worst case, Cholesky decomposition requires $n^3/3$ flops. Solving linear equations requires $2n^2$ flops of forward and backward substitutions. Soft-thresholding operation on $n^2$ variables requires $n^2$ computations. Let $T_1$ be the bound number of iterations, total cost of \textproc{Constrained\_Lasso} is $\mathcal{O} \left(n^3/3 + 3T_1n^2 \right)$. Spectral clustering consists of maximum $\mathcal{O}\left(n^3 \right)$ computations for eigendecomposition and $\mathcal{O}\left(T_2\kappa^2 n \right)$ for \emph{K}-means clustering on $n$ $\kappa$-dimensional eigenvectors, where $T_2$ is the bound number of iterations in \emph{K}-means and $\kappa$ is the number of clusters. The time complexity of SSC-NC is $\mathcal{O}\left(4n^3/3 + 3T_1n^2 + T_2\kappa^2 n\right)$.

{\textbf{CCC}. Similarly to typical clustering methods, CCC computes the correlation matrix which costs $\mathcal{O}\left(n^2 \right)$. Correlation clusterings are afterwards carried out by Adaptive Label Iterated Contitional Modes (AL-ICM) \cite{Besag1986}. AL-ICM, a greedy algorithm, operates in iterative mode. Every fingerprint is initially assigned to a unique label. At each iteration, AL-ICM assigns to a fingerprint the label of its closest fingerprints. This process is repeated until convergence where no label is updated. If $T_3$ is the bound number of itera\-tions, the time complexity of AL-ICM is bounded to $\mathcal{O}\left(T_3n^2 \right)$. In CCC, correlation clustering is performed $Q$ times where $Q$ is the number of similarity thresholding values. Multiple base clusterings are combined to find the final clustering agreement by Weighted Evidence Accumulation Clustering (WEAC) \cite{Huang2015}. The time complexity of WEAC is $\mathcal{O}\left((Q+\log{n})n^2 + Qn \right)$. Finally, $m$ obtained clusters are refined via a merging step which costs $\mathcal{O}\left({m^2\log{m}}\right)$. Total cost of CCC is $\mathcal{O}\left((QT_3 + Q + \log{n} + 1)n^2 + Qn + m^2\log{m}\right)$}. 

\textbf{LS-SSC}. In large-scale contexts, we suppose that RAM can cache only $p$ fingerprints. The dataset is split into $B$ batches, $B=\lceil{\frac{n}{p}}\rceil$. Clustering each batch requires running \textproc{Constrained\_Lasso}, finding \emph{K} nearest neighbors and \textproc{DBSCAN}. Finding \emph{K} nearest neighbors requires sorting each column of sparse representation matrix, which is $\mathcal{O}\left(p^2\log{p} \right)$. In the worst case, DBSCAN visits $p$ points and scans for their neighbors, which costs $\mathcal{O} \left(p^2 \right)$. Total cost of clustering $B$ batches is $\mathcal{O}\left(B\left[ p^3/3 + (3T_1 + \log{p} + 1)p^2\right]\right)$. In our large-scale experiments, recycling step is replicated $B/2$ times on batches of size $p$. Merging and attracting phase work similarly to agglomerative hierarchical clustering, and their time complexity is respectively $\mathcal{O}\left( L^2\log{L} \right)$ and $\mathcal{O} \left(UL\log{U} \right)$, where $L$ is the number of discovered clusters after the first phase and $U$ is the number of unclustered images. Total cost of LS-SSC is $\mathcal{O}\left( \right. 1.5B^2\left[ p^3/3 + (3T_1 + \log{p} + 1)p^2\right]\  + {L^2\log{L}} + UL\log{U} \left. \right)$.

\textbf{Lin-LS}. The time complexity of Lin-LS is analyzed for the first iteration. In the coarse step, the correlation calculation of $B$ batches requires $\mathcal{O}\left(Bp^2 \right)$. If the correlation matrix $n \times n$ has $E$ non-zero entries, Graclus partitioning algorithm \cite{Dhillon2007} has the time complexity of $\mathcal{O}\left(pE/n \right)$. Since the number of clusters in coarse step is fixed to $n^{1/4}$, the calculation of correlation matrix in the fining step costs $\mathcal{O} \left(n^{1/4}b^2 \right)$ where $b$ is the average size of clusters. Markov Clustering Algorithm (MCL) applied on $n^{1/4}$ coarse clusters is bounded to $\mathcal{O} \left( n^{1/4}bK^2 \right)$, where $K$, for abuse of notation, is the maximal number of nonzero entries on each column of the binarized correlation matrix. Similarily to LS-SSC, merging and attraction of Lin-LS can be approximated to $\mathcal{O}\left( L^2\log{L}\right)$ and $\mathcal{O}\left( UL\log{U} \right)$ where $L$ is the discovered number of clusters and $U$ refers to the number of unclustered fingerprints. Since both LS-SSC and Lin-LS aim to obtain high-quality clusters of small size, we can equalize $U,L$ in LS-SSC and $U,L$ in Lin-LS for easy comparison. The first iteration of Lin-LS totally costs $\mathcal{O}\left( {Bp^2 + (K^2b + b^2)n^{1/4} + pE/n  + L^2\log{L} + UL\log{U}}\right)$. The two parameters $E$ and $K$ depend on the cluster distribution in the dataset.

In medium-size datasets where no divide-and-conquer is needed, i.e., $p=n$, SSC-NC and LS-SSC are cubic while Lin-LS and CCC are approximately quadratic. In large-scale datasets, only algorithms designed with divide-and-conquer strategy can be run under the constraint on RAM as well as computational power. The time complexity of LS-SSC is cubic with respect to $p$, while Lin-LS is almost quadratic with respect to $p$ and cluster distribution. In fact, when $n$ becomes very large, we can fix $p$ in LS-SSC as an upper bound, while Lin-LS requires to synthesize the correlation matrix $n \times n$ in the coarse step. Moreover, the time complexity of Lin-LS is analyzed only on the first iteration, the cost of following iterations have to be accounted. Although LS-SSC is cubic with respect to $p$ due to Cholesky decomposition, we optimize this computation by exploiting LAPACK \cite{lapack} whose implementation of Cholesky decomposition is extremely efficient. Our implementation will be made available upon paper acceptance.

%
% EXPERIMENTS
%
\section{Experiments}
\label{sec:EXPERIMENTS}
In this section, we provide experimental analyses of the proposed clustering framework. Based on real data, hyperparameters are selected and used thorough all experiments. We validate the superiority of our method under intensive settings, both on medium and large-scale clustering contexts. 
%
% Experimental settings
%
\subsection{Experimental Settings}
\textbf{Dataset}. All experiments are conducted on JPEG images of Dresden \cite{Gloe2010}  and Vision \cite{Shullani2017}. The top-left regions of size $512 \times 512$ are cropped out for fingerprint extraction. We have tested diverse configurations whose quatitative details are outlined in Table \ref{table:test_config_medium_scale} and Table \ref{table:test_config_large_scale}, considering:
\begin{itemize}
	\item \emph{Cluster symmetry}. On Dresden and Vision, we create symmetric datasets containing $100$ images for each camera, and asymmetric datasets containing all available images on each camera. We denote such configuration on Dresden as $\mathcal{D}^a_c$ $\mathcal{D}^s_c$, and on Vision as $\mathcal{V}^a_c$ $\mathcal{V}^s_c$, where $a$ and $s$ stand for \emph{symmetric} and \emph{asymmetric}, respectively, and $c$ is the number of cameras. 
	\item \emph{Multiple instances of the same model}. On Dresden, we create datasets containing $5$ camera instances of each camera model. Combining with cluster symmetry, we obtain symmetric and asymmetric datasets of this configuration as $\mathcal{D}^{sm}_c$ and $\mathcal{D}^{am}_c$. 
	\item \emph{Number of cameras}. In medium-size datasets, we first select $c=5$, and incrementally add $5$ cameras till $c=20$. 
	\item \emph{Large-scale clustering}. On Dresden, we first select $c=30$, and incrementally add $5$ cameras till the maximum $c=74$, considering all cameras. Since Vision is smaller than Dresden, we start with $c=21$ and incrementally add $3$ cameras till $c=33$. Such configurations on Dresden and Vision are respectively denoted as $\mathcal{LD}^a_c$ and $\mathcal{LV}^a_c$. All these configurations include cameras of same models.
\end{itemize}
\begin{table}
	\caption{Testing configurations on medium-size datasets.}
	\label{table:test_config_medium_scale}
	\scriptsize
	\begin{tabular}{| C{0.65cm} | C{0.65cm}  | C{0.65cm} | C{0.65cm} | C{0.65cm} | C{0.65cm} | C{0.65cm} | C{0.65cm} |}
		\hline
		\multicolumn{2}{| c |}{Configuration}  & \multicolumn{2}{ c |}{\# cameras} & \multicolumn{2}{ c |}{\# models} & \multicolumn{2}{c |}{\# images}\\
		\hline
		Dresden                 & Vision & Dresden & Vision & Dresden & Vision & Dresden & Vision \\
		\hline
		\hline
		$\mathcal{D}^s_5$ & $\mathcal{V}^s_5$ & \multicolumn{2}{c |}{$5$} 	    & \multicolumn{2}{c |}{$5$} 	   & \multicolumn{2}{c |}{$500$} \\
		\hline
		$\mathcal{D}^s_{10}$ & $\mathcal{V}^s_{10}$   & \multicolumn{2}{c |}{$10$} 	& \multicolumn{2}{c |}{$10$} 	  & \multicolumn{2}{c |}{$1000$}  \\
		\hline
		$\mathcal{D}^s_{15}$ & $\mathcal{V}^s_{15}$   & \multicolumn{2}{c |}{$15$} 	& \multicolumn{2}{c |}{$15$} 	  & \multicolumn{2}{c |}{$1500$}  \\
		\hline
		$\mathcal{D}^s_{20}$ & $\mathcal{V}^s_{20}$   & \multicolumn{2}{c |}{$20$} 	& \multicolumn{2}{c |}{$20$} 	  & \multicolumn{2}{c |}{$2000$}  \\
		\hline
		$\mathcal{D}^a_5$ & $\mathcal{V}^a_5$  & \multicolumn{2}{c |}{$5$}  & \multicolumn{2}{c |}{$5$}	& $1089$ & $1041$ \\
		\hline
		$\mathcal{D}^a_{10}$ & $\mathcal{V}^a_{10}$	& \multicolumn{2}{c |}{$10$} 	& \multicolumn{2}{c |}{$10$} 	  & $1954$  & $2110$ \\
		\hline
		$\mathcal{D}^a_{15}$ &	 $\mathcal{V}^a_{15}$	  &  \multicolumn{2}{c |}{$15$}	&  \multicolumn{2}{c |}{$15$} & $3031$ &  $3208$ \\
		\hline
		$\mathcal{D}^a_{20}$ &	 $\mathcal{V}^a_{20}$  	&  \multicolumn{2}{c |}{$20$}	&  \multicolumn{2}{c |}{$20$}	 & $4186$  & $4435$ \\
		\hline
		$\mathcal{D}^{sm}_{5}$  & $-$    & $5$     & $-$    & $1$     & $-$    & $500$   & $-$    \\
		\hline
		$\mathcal{D}^{sm}_{10}$ & $-$    & $10$    & $-$    & $2$     & $-$    & $1000$  & $-$    \\
		\hline
		$\mathcal{D}^{sm}_{15}$ & $-$    & $15$    & $-$    & $3$     & $-$    & $1500$  & $-$    \\
		\hline
		$\mathcal{D}^{sm}_{20}$ & $-$    & $20$    & $-$    & $4$     & $-$    & $2000$  & $-$    \\
		\hline
		$\mathcal{D}^{am}_{5}$  & $-$    & $5$     & $-$    & $1$     & $-$    & $855$   & $-$    \\
		\hline
		$\mathcal{D}^{am}_{10}$ & $-$    & $10$    & $-$    & $2$     & $-$    & $2663$  & $-$    \\
		\hline
		$\mathcal{D}^{am}_{15}$ & $-$    & $15$    & $-$    & $3$     & $-$    & $3558$  & $-$    \\
		\hline
		$\mathcal{D}^{am}_{20}$ & $-$    & $20$    & $-$    & $4$     & $-$    & $4540$  & $-$    \\
		\hline
	\end{tabular}
\end{table}

\begin{table}
	\centering
	\caption{Testing configurations on large-scale datasets.}
	\label{table:test_config_large_scale}
	\scriptsize
	\begin{tabular}{| C{0.65cm} | C{0.65cm}  | C{0.65cm} | C{0.65cm} | C{0.65cm} | C{0.65cm} |}
		\hline
		\multicolumn{2}{| c |}{Configuration}  & \multicolumn{2}{ c |}{\# cameras} & \multicolumn{2}{c |}{\# images}\\
		\hline
		Dresden               & Vision                & Dresden & Vision & Dresden & Vision \\
		\hline					
		$\mathcal{LD}^a_{30}$ & $\mathcal{LV}^a_{21}$ & $30$    & $21$   & $6596$  & $4397$ \\
		\hline
		$\mathcal{LD}^a_{35}$ & $\mathcal{LV}^a_{24}$ & $35$    & $24$   & $7538$  & $5051$ \\
		\hline
		$\mathcal{LD}^a_{40}$ & $\mathcal{LV}^a_{27}$ & $40$    & $27$   & $8545$  & $5773$ \\
		\hline
		$\mathcal{LD}^a_{45}$ & $\mathcal{LV}^a_{30}$ & $45$    & $30$   & $9635$  & $6377$ \\
		\hline
		$\mathcal{LD}^a_{50}$ & $\mathcal{LV}^a_{33}$ & $50$    & $33$   & $10765$ & $7070$ \\
		\hline
		$\mathcal{LD}^a_{55}$ & $-$                   & $55$    & $-$    & $11673$ & $-$    \\
		\hline
		$\mathcal{LD}^a_{60}$ & $-$                   & $60$    & $-$    & $12729$ & $-$    \\
		\hline
		$\mathcal{LD}^a_{65}$ & $-$                   & $65$    & $-$    & $13995$ & $-$    \\
		\hline
		$\mathcal{LD}^a_{70}$ & $-$                   & $70$    & $-$    & $14915$ & $-$    \\
		\hline
		$\mathcal{LD}^a_{74}$ & $-$                   & $74$    & $-$    & $15677$ & $-$    \\
		\hline
	\end{tabular}
\end{table}
		
\textbf{Performance metric.} We report performance in $\mathcal{F}$-measure and Adjusted Rand Index (ARI). In the presence of outliers (unclustered fingerprints), we follow \cite{Lin2017} and treat outliers differently in the computation of True Positive ($\overline{TP}$) and False Positive ($\overline{TP}$). Specifically, 
\begin{itemize}
	\item True Positive ($\overline{TP}$): the number of image pairs from the same cluster which are assigned to the same cluster, \emph{excluding outliers}.
	\item False Positive ($\overline{FP}$): the number of image pairs from different clusters which are assigned to the same cluster, \emph{excluding outliers}.
	\item True Negative ($TN$): number of image pairs from different clusters which are assigned to different clusters.
	\item False Negative ($FN$): number of image pairs from the same cluster which are assigned to different clusters.
\end{itemize}
$\mathcal{F}$-measure is computed based on precision ($\mathcal{P}$) and recall ($\mathcal{R}$):
	
\begin{IEEEeqnarray}{rCl}
	\mathcal{P} &=& \frac{\overline{TP}}{\overline{TP} + \overline{FP}} \text{,} \quad \mathcal{R} = \frac{\overline{TP}}{\overline{TP} + FN} \text{,} \quad \mathcal{F} = 2 \cdot \frac{\mathcal{P} \cdot \mathcal{C}}{\mathcal{P} + \mathcal{C}} \text{.} \IEEEnonumber
\end{IEEEeqnarray}
Rand Index (RI) and ARI are computed as:
\begin{IEEEeqnarray}{rCl}
	\text{RI} &=& \frac{\overline{TP} + TN}{\overline{TP} + TN + \overline{FP} + FN} \text{,} \quad
	\text{ARI} = \frac{\text{RI} - \mathbf{E}[\text{RI}]}{1 - \mathbf{E}[\text{RI}]} \text{,}\IEEEnonumber
\end{IEEEeqnarray}
where $\mathbf{E}[\text{RI}]$ is the expected value of RI and is computed based on the expected value of $\overline{TP}$ and $TN$.
\begin{IEEEeqnarray}{rCl}
	\mathbf{E}[\text{RI}] &=& \frac{\mathbf{E}[\overline{TP}] + \mathbf{E}[TN]}{\overline{TP} + TN + \overline{FP} + FN} \IEEEnonumber \text{.}
\end{IEEEeqnarray}
	
The readers can refer to \cite{Hubert1985} for more details of ARI computation. When the number of outliers is zero, $\mathcal{F}$-measure and ARI become canonically defined.

For comparing the number of clusters discovered by each algorithm, we follow \cite{Lin2017} to report the ratio $L_p/L_g$ where $L_p$ refers to the number of predicted clusters and $L_g$ the number of ground-truth clusters. Differently to \cite{Lin2017} where $L_p$ only accounts for \emph{unique} predicted clusters, i.e., $L_p \leq L_g$, it is possible in our evaluation that $L_p/L_g > 1$ if an algorithm overestimates, or $L_p/L_g < 1$ if under-estimating the number of ground-truth clusters.
	
\textbf{Performance comparison}. We compare the results of the proposed methodologies with the state of the art. Tests have been done also with hierarchical clustering \cite{GarciaVillalba2015}, Markov Random Field \cite{Li2010_2}, and Spectral Clustering with Normalized Cut criterion \cite{Amerini2014}, but for the sake of space and readability we only present comparisons with the following top performing works:

\begin{itemize}
	\item \emph{Multiclass Spectral Clustering (MSC)} \cite{Liu2010}. A star graph is built with $5$ nearest neighbors, as suggested in \cite{Liu2010}.
	\item \emph{Lin's Large-Scale (Lin-LS) method} \cite{Lin2017}. Lin-LS is implemented with all parameters recommended from \cite{Lin2017}: compressed fingerprints ($256 \times 256$) are binarized by threshold $t_b = 0.008$, while original-size fingerprints ($1024 \times 1024$) are binarized by threshold $t_b = 0.005$. In order to take divide-and-conquer strategy into effect on medium-size datasets, each dataset is split into two equal batches and only one is loaded at once.
	\item \emph{Correlation Clustering with Consensus (CCC)} \cite{Marra2017}. Results of CCC are acquired from the implementation provided by the authors. No parameter needs to be specified.
	\item \emph{Sparse Subspace Clustering (SSC)} \cite{Phan2017}. SSC is implemented similarly to SSC-NC but without the non-negativity constraint.
\end{itemize}

We analyze the performance of SSC-NC to see the effectiveness of non-negativity constraint, and LS-SSC to verify its adaptation on medium-size and large-scale datasets. To simulate divide-and-conquer on medium-size datasets, LS-SSC splits each dataset into two equal batches and only one is loaded at once in the same manner as Lin-LS. 

Under large-scale datasets, LS-SSC is compared only to Lin-LS since these methods are particularly designed for large-scale contexts. One matter of clustering on large-scale datasets is the lack of memory. Since only a limited number of fingerprints can be allocated on RAM, we fix this bound to $4000$ ($\approx 4$ GBs are required to store fingerprints).

Due to some randomization used in MSC, CCC, Lin-LS and LS-SSC, those methods are run $10$ times, and the average scores are reported.
%
% Parameter selections
%
\subsection{Hyperparameter Selection}
\label{sec:hyperparameter_selection}
In order to select a number of parameters required by our methodologies we collect a dataset, obviously different from the test one.
From RAISE dataset \cite{Dang-Nguyen2015} we extract $200$ raw images from Nikon D90 and $250$ from D7000, and perform JPEG compression (quality factor $98$). Since there are only $76$ raw images of Nikon D40, we leave them out and instead select $300$ JPEG images (default JPEG quality setting) from an external Canon 600D. We refer to this dataset as $\mathcal{D}_{\text{dev}}$ including $750$ images from $3$ cameras.

\textbf{Selecting $\eta$}. $\eta$ is the augmented Lagrangian hyperparameter which stands for how much penalty added in order to enforce the equality $\mathbf{Z=V}$. This parameter partially decides the convergence speed of \textproc{Constrained\_Lasso}. Small $\eta$ means slow convergence but with high accurate solutions, while large $\eta$ accelerates convergence speed but results in modest accurate solutions. Since sparse representation learning is followed by a clustering procedure, solutions with modest accuracy are sufficient. On $\mathcal{D}_{\text{dev}}$, ${\eta \in [1.0,1.3]}$ results in acceptable solutions and fast convergence. We adopt ${\eta=1.0}$ in all experiments.

\textbf{Selecting $\gamma$}. On $\mathcal{D}_{\text{dev}}$, we vary $\gamma$ in the range $[0.0001, 0.02]$ and select $\gamma = 0.0018$ that minimizes the cost function defined in  \cite{Phan2017} taken into account normalized cuts and eigengaps as criterions.

\begin{figure}
	\centering
	\includegraphics[width=\linewidth]{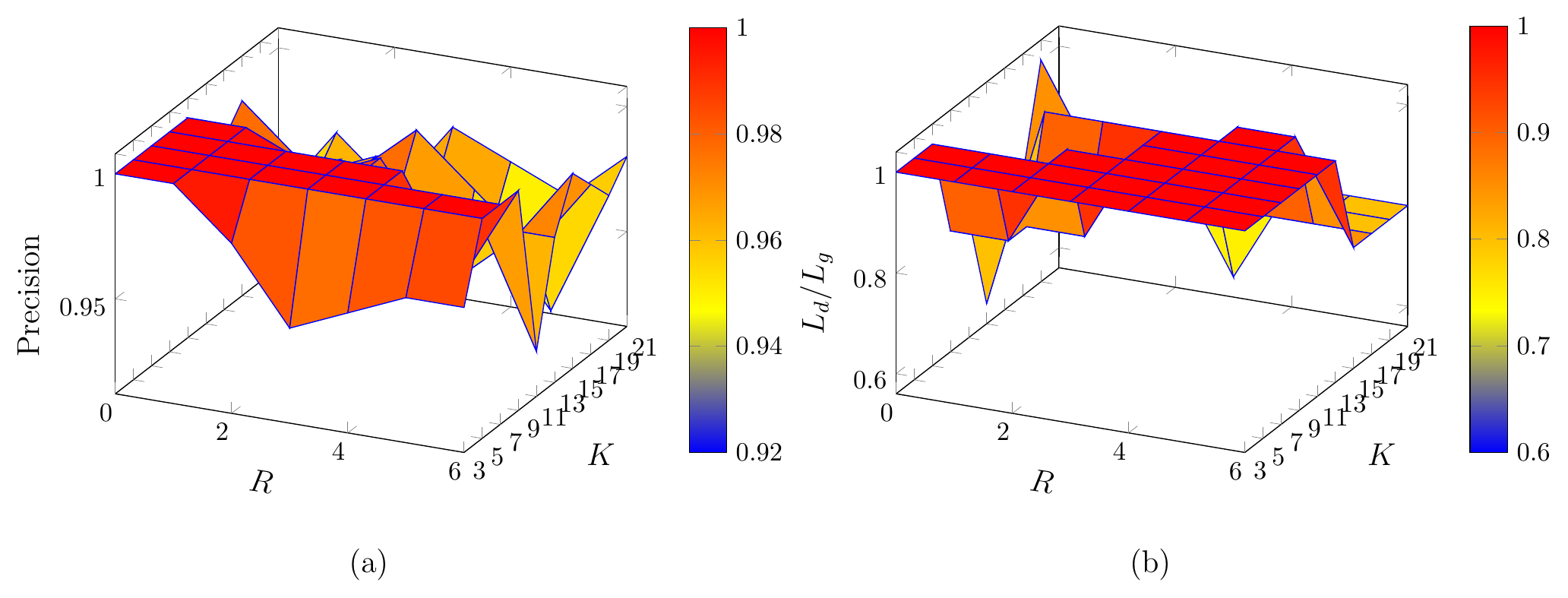}
	\caption{Precision and $L_d/L_g$ with respect to diverse values of $K$ and $\#$ recycling steps.}
	\label{fig:para_setting}
\end{figure}

\textbf{Jointly selecting $R$ and $K$}. In LS-SSC, the main goal of recycling is to reduce the number of undiscovered ground-truth clusters. Let us denote as $L_d, L_g$ the number of ground-truth clusters discovered after merging phase and the number of ground-truth clusters, respectively. The strategy is to adopt the number of recycling steps $R$ such that ${L_d/L_g \rightarrow 1}$ and discovered clusters are pure, namely ${\text{Precision} \rightarrow 1}$. Another parameter which impacts on $L_d$ is the number of nearest neighbors $K$. Small $K$ means more ground-truth clusters are likely to be discovered, otherwise only noticeably dense clusters are discovered. We conduct experiments on an asymmetric dataset from Dresden containing $5$ cameras coming from different models. We split the dataset into $6$ equal batches of size $\approx 182$ in order to simulate splitting step. Figure \ref{fig:para_setting} depicts precision of discovered clusters after merging step in panel (a), and the ratio $L_d/L_g$ in panel (b). It is clear that ${K = 5}$ is a reasonable choice for discovering pure ground-truth clusters. From these plots one can argue that selecting $R=0$ allows to obtain the highest precision in this case. However, it is important to remember that recycling plays an important role since it helps discover more hidden clusters. In principle, high value of $R$ should be chosen considering the computational complexity, but the precision is likely to drop if we run more recycling steps with big $K$. In large-scale contexts, we adopt $R = \lfloor B/2 \rfloor$, where $B$ is the number of batches. In medium-scale contexts, where computational requirement is less important, we run recycling until there is no noticeable subclusters discovered.
%
% Numeric results on medium-size datasets
%
\begin{figure*}[h]
	\centering
	\captionsetup{justification=centering}
	\includegraphics[width=0.85\linewidth]{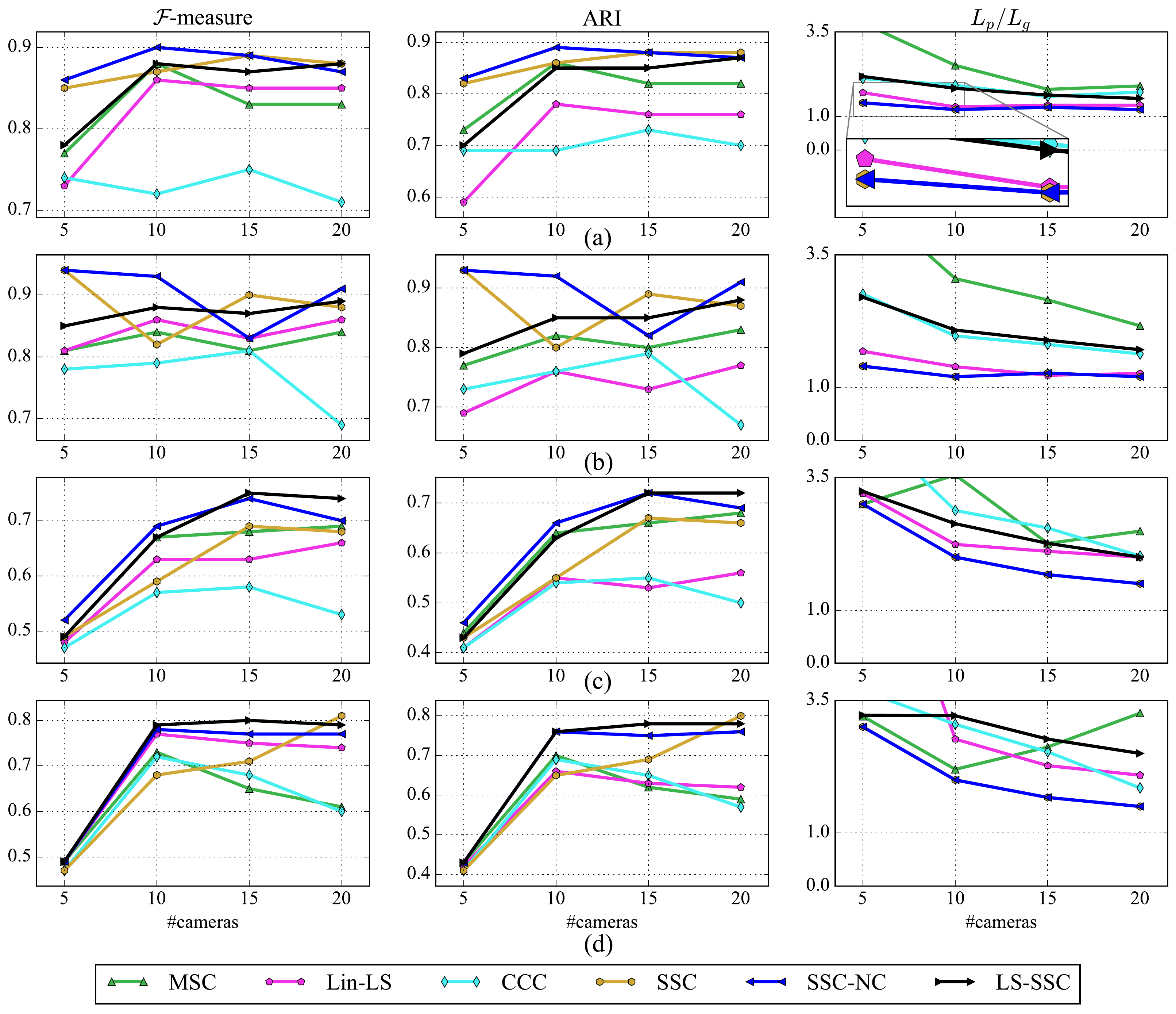}
	\caption{Clustering performance on medium-size datasets of Dresden: (a) symmetric: $\mathcal{D}^s_5, \mathcal{D}^s_{10}, \mathcal{D}^s_{15}, \mathcal{D}^s_{20}$ (b) asymmetric: $\mathcal{D}^a_5, \mathcal{D}^a_{10}, \mathcal{D}^a_{15}, \mathcal{D}^a_{20}$ (c) symmetric + same model: $\mathcal{D}^{sm}_5, \mathcal{D}^{sm}_{10}, \mathcal{D}^{sm}_{15}, \mathcal{D}^{sm}_{20}$ (d) asymmetric + same model:  $\mathcal{D}^{am}_5, \mathcal{D}^{am}_{10}, \mathcal{D}^{am}_{15}, \mathcal{D}^{am}_{20}$. Better viewed in color.}
	\label{fig:results_medium_dresden}
\end{figure*}
\begin{figure}[!t]
	\centering
	\captionsetup{justification=centering}
	\includegraphics[width=\linewidth]{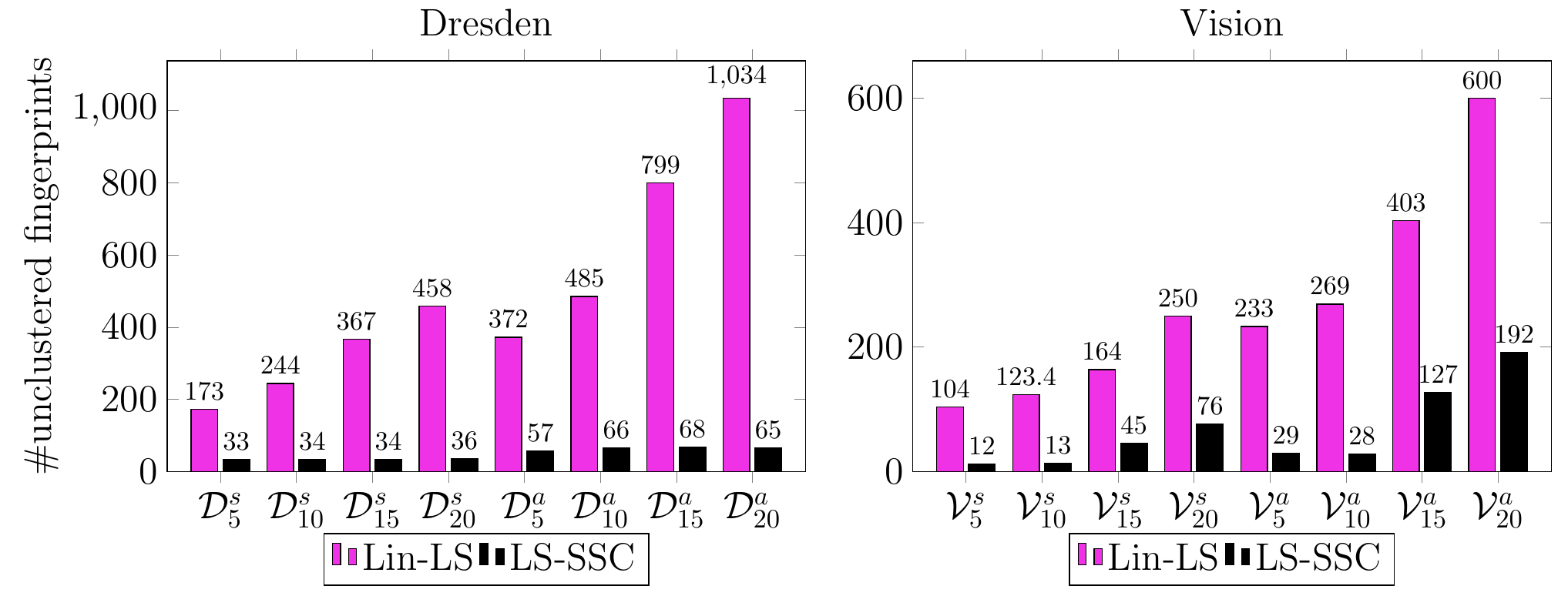}
	\caption{Number of unclustered fingerprints on medium-size datasets of Dresden and Vision.}
	\label{fig:unclustered_fingerprints_medium}
\end{figure}
\subsection{Numeric Results on Medium-size Datasets}

\begin{figure*}[!t]
	\centering
	\captionsetup{justification=centering}
	\includegraphics[width=0.85\linewidth]{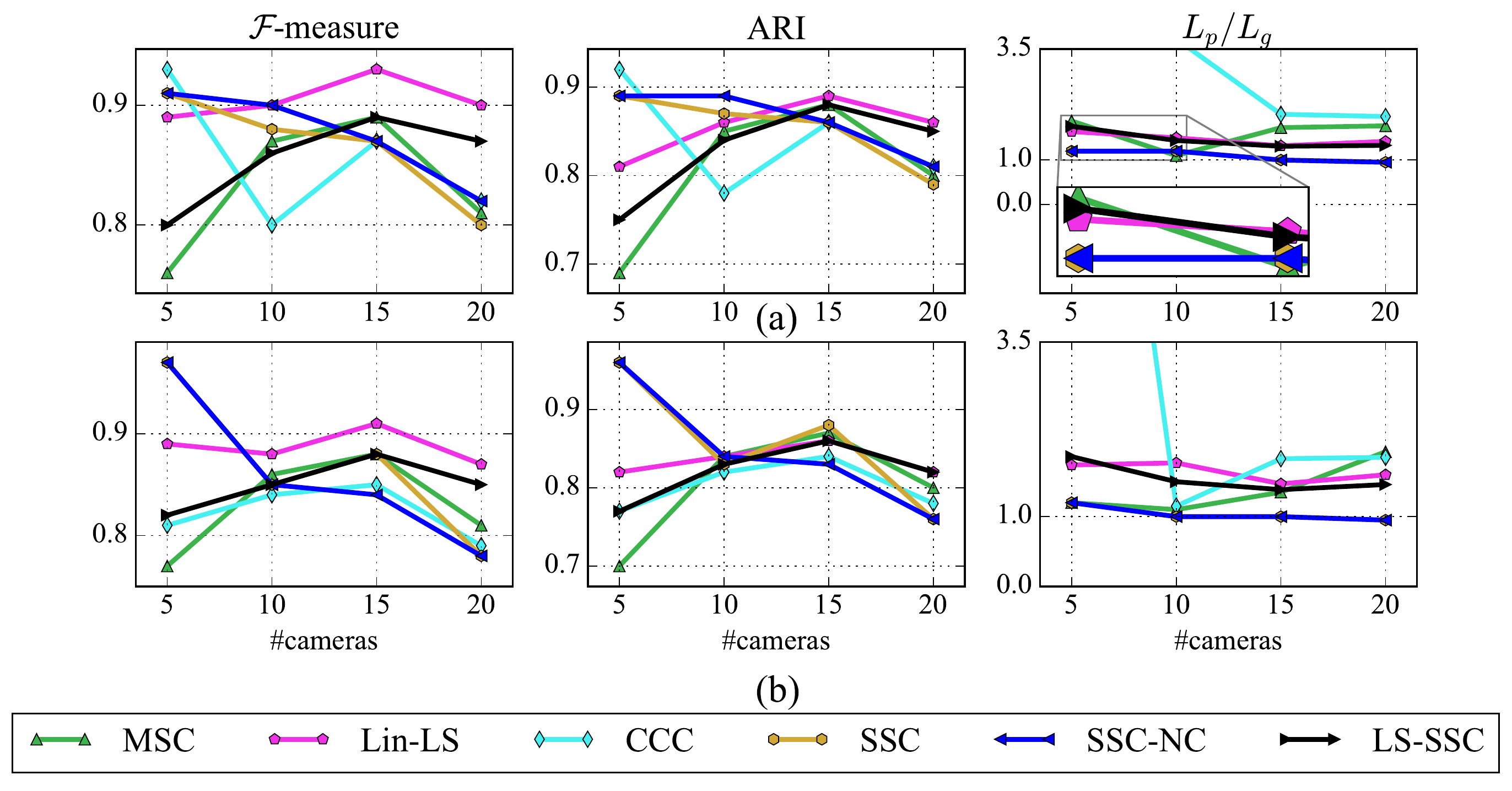}
	\caption{Clustering performance on medium-size datasets of Vision: (a) symmetric: $\mathcal{V}^s_5, \mathcal{V}^s_{10}, \mathcal{V}^s_{15}, \mathcal{V}^s_{20}$ (b) asymmetric: $\mathcal{V}^a_5, \mathcal{V}^a_{10}, \mathcal{V}^a_{15}, \mathcal{V}^a_{20}$. Better viewed in color.}
	\label{fig:results_medium_vision}
\end{figure*}

We report performance of all methods on medium-size datasets with the maximum number of images ranging from $4000$ to $5000$. 

Results on Dresden suggest that MSC performs relatively well on symmetric (in Figure \ref{fig:results_medium_dresden} (a)) and asymmetric (in Figure \ref{fig:results_medium_dresden} (b)) datasets. MSC applies an extra step before clustering. It is the creation of a star graph among fingerprints, where noisy connections are partially eliminated. The star graph can be considered as a suboptimal sparse representation matrix of data. Differently to MSC, SSC finds a sparse representation of data by solving an optimization problem. In Figure \ref{fig:results_medium_dresden}, SSC outperforms MSC in most configurations with high $\mathcal{F}$-measure. As an improved version of SSC, SSC-NC performs equally or better than SSC in the majori\-ty of symmetric and asymmetric datasets. Balanced precision and recall are obtained, gaining high $\mathcal{F}$-measure. The number of predicted clusters $L_p$ obtained by SSC and SSC-NC are identical,  approximating well the number of ground-truth clusters $L_g$. Such approximation is the best among all tested algorithms.

Although Lin-LS and LS-SSC are especially designed for large-scale datasets, they produce convincing results also on medium-size datasets. Lin-LS aims to obtain high-quality clusters of small size, resulting in high precision. Comparing to Lin-LS, LS-SSC obtains less precise clusters but the precision is still high without penalizing recall. Thanks to this balanced behavior, LS-SSC outperforms Lin-LS in terms of $\mathcal{F}$-measure and ARI. To keep precision high, both Lin-LS and LS-SSC tend to overestimate the number clusters in medium-size datasets. 
\begin{figure*}[!t]
	\centering
	\captionsetup{justification=centering}
	\includegraphics[width=\linewidth]{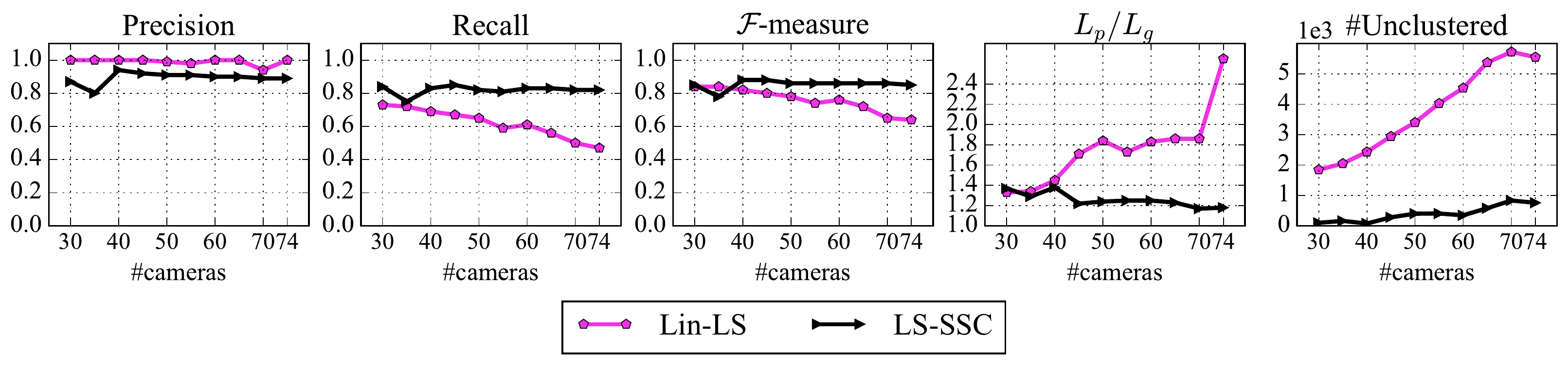}
	\caption{Clustering results on large-scale datasets of Dresden.}
	\label{fig:results_ls_dresden}
\end{figure*}

\begin{figure*}[!t]
	\centering
	\captionsetup{justification=centering}
	\includegraphics[width=\linewidth]{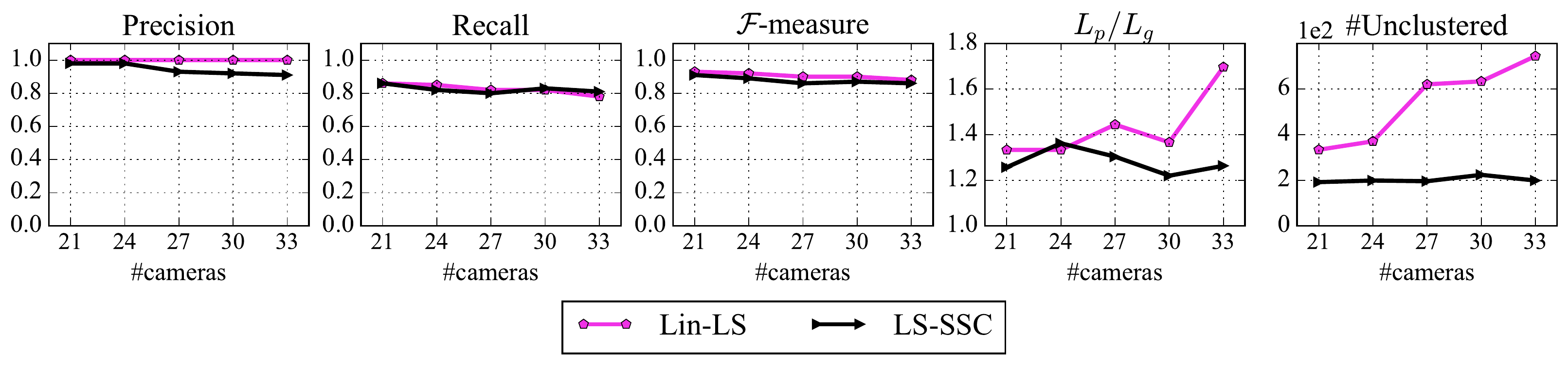}
	\caption{Clustering results on large-scale datasets of Vision.}
	\label{fig:results_ls_vision}
\end{figure*}

Zooming into the cases where cameras of the same model share some commonalities in SPNs, this clearly introduces a certain level of ambiguity. In Figure \ref{fig:results_medium_dresden} (c) and (d) we report results on datasets containing multiple camera models, each model with $5$ camera instances. Despite the fact that all methods suffer from performance degradation, SSC-NC outperforms other methods in $\mathcal{D}_{5}^{sm}, \mathcal{D}_{10}^{sm}$, while LS-SSC is superior in all other configurations. In Figure \ref{fig:results_medium_dresden} (c) and (d), the superiority of SSC-NC over SSC is evident. We argue that, in such complicated contexts where SPNs of the same camera model stay close to each other, SSC-NC can find a better representations of data.

We replicate the evaluation of all methods on medium-size datasets of Vision, see Figure \ref{fig:results_medium_vision}. MSC, SSC-NC and LS-SSC perform on par with each other, but SSC-NC achieves more accurate estimation on the number of clusters. On the other hand, SSC-NC also obtains more accurate results than SSC in almost all configurations ($7$ out of $8$). It seems that Lin-LS outperforms all other methods, however, we argue that its performance gain is partially due to high number of unclustered fingerprints it produces. We show in Figure \ref{fig:unclustered_fingerprints_medium} the number of unclustered fingerprints of LS-SSC and Lin-LS on medium-size datasets of Dresden and Vision. It is evident that Lin-LS produces more outliers than LS-SSC, thus gaining a certain advantage over precision, and then $\mathcal{F}$-measure as a consequence.
%
% Numeric results on large-scale datasets
%
\subsection{Numeric Results on Large-scale Datasets}
In practice, there exist large-scale contexts where a large number of images need to be clustered. In Dresden, we conduct experiments on datasets containing $30$ to $74$ cameras, and the number of images exceeds $6000$, while in Vision the number of cameras ranges from $21$ to $33$ and the number of images exceeds $4000$. To the best of our knowledge, Lin-LS \cite{Lin2017} is the only method proposed for large-scale clustering of camera fingerprints, thus results are compared only with it. 

As depicted in Figure \ref{fig:results_ls_dresden}, Lin-LS achieves high precision, which means $\overline{FP}$ is negligible. Nevertheless, in order to keep high precision a noticeable number of fingerprints are not clustered. Unclustered fingerprints essentially causes low recall, or equivalently high $FN$ due to the separation of pairs belonging to the same cluster. On the contrary, LS-SSC produces less precise clusters with precision from $80 \%$ to $100 \%$. One advantage of our method is the achievement of relatively high recall which slightly oscillates around $80\%$. Apart from keeping precision and recall balanced, we obtain high $\mathcal{F}$-measure. LS-SSC can cluster the whole Dresden dataset with $\mathcal{F}$-measure higher than $80\%$ which substantially improves the $64 \%$ obtained by Lin-LS. The improvement of LS-SSC over Lin-LS should be further amplified because Lin-LS requires to access $1024 \times 1024$ fingerprints in refining step while LS-SSC only works on $512 \times 512$ fingerprints. Moreover, as depicted in Figure \ref{fig:results_ls_dresden} (last panel), LS-SSC produces a higher number of clusters than the ground-truth clusters, but the ratio between the two quantities is relatively constant when the dataset size grows. Vice versa, for Lin-LS this ratio rapidly increases.
	
Shown in Figure \ref{fig:results_ls_vision} are the performance of Lin-LS and LS-SSC on Vision dataset. Lin-LS again produces highly precise clusters, but tends to overestimate the number of ground-truth clusters. The $\mathcal{F}$-measure scores of the two methods are close since unclustered fingerprints are not accounted for precision computation.

In Lin-LS, the main cause of unclustered fingerprints are due to the merging step. If the merging threshold is too high, small subclusters cannot be merged to form larger subclusters, and thus filtered out in the end. On the other hand, in LS-SSC a fingerprint is unclustered if the correlation between fingerprint and all available cluster centroids is smaller than a threshold that was used to exclude the null hypothesis. Also for the case of large-scale datasets, we show the number of unclustered fingerprints in Lin-LS and LS-SSC, see last panel of Figure \ref{fig:results_ls_dresden}, \ref{fig:results_ls_vision}. In this scenario, it is clear that to keep precision high Lin-LS produces large number of unclustered fingerprints, not comparable with unclustered fingerprints in LS-SSC. The advantage of this mechanism is to reduce false alarm rate, but its downside is evident since data of interest could be ignored by the algorithm. LS-SSC provides a reasonable tradeoff allowing to cluster large-scale databases without skipping too many images which might be important for forensic analysis.

\subsection{LS-SSC Robustness Analysis}
In this section, we analyze the robustness of LS-SSC in more realistic testing configurations. 

\subsubsection{Presence of outliers}
Firstly, we test the robustness of LS-SSC to outliers. We select images coming from $20$ cameras of Vision, and add $50$ images randomly collected from Facebook (from different entities) to make sure that they do not share the same source camera. On this dataset, LS-SSC achieves $\mathcal{F}$-measure $0.89$. Remarkably, LS-SSC assigns $69$ images as unclustered, in which $33$ out of $50$ images are truthfully outliers.

\subsubsection{Double JPEG compression}
Images taken via smartphones usually undergo double JPEG compression once being available on social media sites. Therefore, we test the robustness of LS-SSC on images coming from $20$ cameras of Vision, further compressed using \texttt{convert} tool provided by \texttt{ImageMagick}. The compression quality ranges from $50$ to $95$ (step $5$). Results in Table \ref{table:num_results_vision_20_compressed} expose very reasonable and pretty stable performance of LS-SSC over different quality factors. Indeed, the algorithm is generally robust to double JPEG compression if the quality factor of the second compression is more than $65$. Clustering performance starts to drop if images are aggressively compressed (quality factor smaller than $65$).

\begin{table}[!h]
	\centering
	\scriptsize
	\caption{Numeric results of LS-SSC on double compressed images.}
	\begin{tabular}{ C{0.5cm} | C{0.35cm} C{0.35cm} C{0.35cm} C{0.35cm} C{0.35cm} C{0.35cm} C{0.35cm} C{0.35cm} C{0.35cm} C{0.35cm} }
		\multirow{2}{*}{Metric} & \multicolumn{10}{c}{Quality factor} \\
		\cline{2-11}
             & 50     & 55     & 60     & 65     & 70     & 75     & 80     & 85     & 90     & 95     \\
		\hline
		$\mathcal{P}$ & $0.79$ & $0.81$ & $0.84$ & $0.88$ & $0.90$ & $0.93$ & $0.94$ & $0.95$ & $0.93$ & $0.96$ \\
		$\mathcal{R}$ & $0.58$ & $0.61$ & $0.66$ & $0.72$ & $0.77$ & $0.80$ & $0.83$ & $0.86$ & $0.85$ & $0.88$ \\
		$\mathcal{F}$ & $0.67$ & $0.70$ & $0.74$ & $0.79$ & $0.83$ & $0.86$ & $0.88$ & $0.90$ & $0.89$ & $0.92$ \\
		ARI           & $0.65$ & $0.68$ & $0.73$ & $0.78$ & $0.82$ & $0.85$ & $0.87$ & $0.89$ & $0.88$ & $0.88$ \\
	\end{tabular}
	\label{table:num_results_vision_20_compressed}
\end{table}
	
In practice, images may come from online social networks where they  undergo double compression. In such scenario, SPNs are further distorted due to resizing, and it has been confirmed by \cite{Marra2017} that performance of all methods drop.
	
	\subsubsection{Different SPN sizes}
	Next, we validate the robustness of LS-SSC to different sizes of SPN. We pick the same set of images used in previous experiment, but crop the top-left region to $4$ different sizes: $256\times256, 512\times512, 768\times768, 1024\times1024$. The hyperparameter $\gamma$ is also re-estimated on the development set where images are cropped to similar sizes. The values of $\gamma$ for each of corresponding size are $0.0045, 0.0018, 0.0012, 0.0008$. In Table \ref{table:num_results_vision_20_size}, the performance generally improves if larger-size SPNs are used. Nevertheless, the results also suggest that using SPN sizes larger than $512 \times 512$ is not the key for the success of LS-SSC. Indeed, using $768 \times 768$ does not gain any improvement over $512 \times 512$ SPNs, and using $1024 \times 1024$ SPNs brings only a minor improvement.
	
	\begin{table}[!h]
		\centering
		\scriptsize
		\caption{Numeric results of LS-SSC on SPNs of different sizes.}
		\begin{tabular}{ C{0.5cm} | C{1.5cm} C{1.5cm} C{1.5cm} C{1.5cm}}
			\multirow{2}{*}{Metric} & \multicolumn{4}{c}{SPN size} \\
			\cline{2-5}
			              & $256 \times 256$ & $512 \times 512$ & $768 \times 768$ & $1024 \times 1024$ \\
			\hline
			$\mathcal{P}$ & $0.85$           & $0.92$           & $0.92$           & $0.89$             \\
			$\mathcal{R}$ & $0.84$           & $0.84$           & $0.85$           & $0.90$             \\
			$\mathcal{F}$ & $0.84$           & $0.88$           & $0.88$           & $0.89$             \\
			ARI           & $0.83$           & $0.87$           & $0.87$           & $0.88$             \\
		\end{tabular}
		\label{table:num_results_vision_20_size}
	\end{table}
	
	\subsubsection{Few images per camera}
	In some specific contexts, forensic analysts might face with databases where the number of cameras is higher than the average number of images acquired by each camera (one camera per each model). To simulate such context, we start with an original set of $20$ cameras selected from Dresden. The number of images on each camera alternatively ranges from $10$ to $50$ (step $10$). For each image, we crop at $50$ different positions, ending an augmented set of images coming from $50$ cameras. Finally we obtain a dataset of $12000$ images of $400$ cameras. It is a challenging dataset since the number of cameras is high, while the number of images for each camera is much lower. LS-SSC assigns images into $258$ clusters, and $469$ images remain unclustered. Obviously, many small-size clusters are hard to be discovered due to random splitting.  
	
	It is acknowledged in \cite{Lin2017} that Lin-LS is especially designed to cope with such scenarios. However, such capability comes at a cost of discarding many outliers, which might leave images of interest out of consideration. Lin-LS assigns images into $892$ clusters, while $4083$ images remain unclustered. We obtain an $\mathcal{F}$-measure $47\%$ in this dataset, while Lin-LS achieves $52\%$, at a price of a $10$ times larger number of unclustered images.
	
	In this scenario, LS-SSC performs not very well, but this is somehow inherently defined in the method itself. Indeed, we know from the theory that learning sparse representation of camera fingerprints requires a sufficient number of images per camera. Without this assumption, the algorithm might learn inexact representations which usually result in high $\overline{FP}$.
	
	\subsection{Running time analysis}
	
\begin{figure}[!t]
		\centering
		\captionsetup{justification=centering}
		\includegraphics[width=0.8\linewidth]{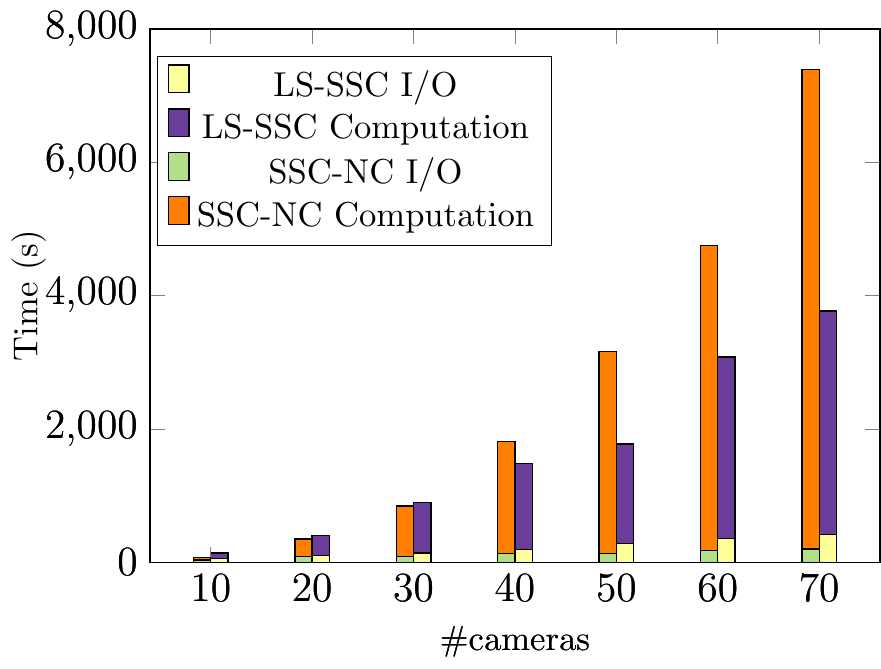}
		\caption{Running time of SSC-NC and LS-SSC.}
		\label{fig:running_time_analysis}
	\end{figure}

	We measure the running time of SSC-NC and LS-SSC on Dresden images, where the number of cameras ranges from $10$ to $70$. To observe the running time of SSC-NC we assume RAM is sufficient to catch all fingerprints of $70$ cameras, and allows to solve the optimization in Eq. (\ref{proposed_opt}). 
	Figure \ref{fig:running_time_analysis} reveals the fact that LS-SSC requires higher I/O cost due to extra reading/writing operations. SSC-NC, on the other hand, requires much higher computational cost, which are critical in practical usages. For LS-SSC, it takes approximately $1$ hour and $20$ minutes to cluster the whole Dresden dataset. In the case of limited RAM, LS-SSC requires more I/O time while SSC-NC cannot be operated.
	
	%
	% CONCLUSION
	%
	\section{Conclusion}
	We have introduced a clustering framework by exploiting linear dependencies among SPNs in their intrinsic vector subspaces. Each SPN is expressed as a sparse linear combination of all other SPNs. Finding such sparse combinations is equivalent to solving LASSO with constraints, which is done efficiently by ADMM method. Our algorithm can be extended to the case of large-scale databases thanks to the proposed divide-and-conquer strategy. Experiments prove the advantage of sparse representation over normalized correlation.
	
	Future extensions will be dedicated to combining sparse representation learning and clustering into a unified end-to-end procedure. Moreover, we foresee to further study the impact of cluster cardinality on the method performance.
	
	\appendices
	
	\section{Derivation of $\V$ update in Algorithm \ref{alg:lasso_solving_admm}} 
	\label{app:v_update}
	At each iteration of \textproc{CONSTRAINED\_LASSO}, $\V$ is updated by:
	\begin{IEEEeqnarray}{rCl}
		\V &=& \arg \underset{\V}{\min} \; f(\V) \text{,} \IEEEnonumber
	\end{IEEEeqnarray}
	where
	\begin{IEEEeqnarray}{rCl}
		f(\V)&=& \gamma\| \V\|_1 + \langle \bLambda,\Z-\V \rangle + \frac{\eta}{2} \|\Z - \V\|_F^2 \text{.} \IEEEnonumber \\
		\frac{\partial f}{\partial \V_{ij}} &=& \gamma\frac{\partial |\V_{ij}|}{\partial \V_{ij}} + \eta\V_{ij} - \eta \left( \Z_{ij} + \frac{\bLambda_{ij}}{\eta} \right)  \IEEEnonumber \\
		&=& \begin{cases}
		-\gamma + \eta\V_{ij} - \eta \left( \Z_{ij}+\frac{\bLambda_{ij}}{\eta} \right), \quad &\V_{ij} < 0 \IEEEnonumber \\
		\gamma + \eta\V_{ij} - \eta \left( \Z_{ij}+\frac{\bLambda_{ij}}{\eta} \right), \quad &\V_{ij} > 0 \IEEEnonumber \\
		\text{undefined,} \quad &\V_{ij}=0
		\end{cases} \text{.} \IEEEnonumber \\
		\frac{\partial f}{\partial \V_{ij}} &=& 0 \; \text{then} \IEEEnonumber  \\
		\V_{ij} &=& S_{\frac{\gamma}{\eta}}\left( \Z_{ij} + \frac{\bLambda_{ij}}{\eta} \right) \IEEEnonumber \\
		&=& \begin{cases}
		\frac{\gamma}{\eta} + \Z_{ij} + \frac{\bLambda_{ij}}{\eta} \text{,} &\text{if} \quad \Z_{ij} + \frac{\bLambda_{ij}}{\eta} < -\frac{\gamma}{\eta} \IEEEnonumber \\
		\frac{-\gamma}{\eta} +  \Z_{ij} + \frac{\bLambda_{ij}}{\eta} \text{,} &\text{if} \quad \Z_{ij} + \frac{\bLambda_{ij}}{\eta} > \frac{\gamma}{\eta} \IEEEnonumber \\
		0 \text{,} &\text{if} \quad \left| \Z_{ij} + \frac{\bLambda_{ij}}{\eta} \right| \leq \frac{\gamma}{\eta}
		\end{cases} \text{.}
	\end{IEEEeqnarray} 
	Solution $\V \in \mathbb{R}^{n \times n}$ might violate two constraints in Eq. (\ref{proposed_opt}). Denote $\mathcal{C}_1$ the set of all zero-diagonal and $\mathcal{C}_2$ the set of non-negative matrices. $\mathcal{C}_1, \mathcal{C}_2$ are convex. To impose the two constraints on $\V$, it is equivalent to find $\V^{(1,2)} \in \mathcal{C}_1 \cap \mathcal{C}_2$ that minimizes $f$. This can be obtained via von Neumann's alternating projections \cite{Cheney1959}: first Euclidean projection onto $\mathcal{C}_1$, and second Euclidean projection onto $\mathcal{C}_2$. Since $\V \in \mathbb{R}^{n \times n}$ is a minimizer of $f$, $\V^{(1,2)}$ can be obtained by two successive projections:
	\begin{IEEEeqnarray}{rlll}
		\V^{(1)} = \text{arg} \underset{\tilde{\V} \in \mathcal{C}_1}{\min} \| \V - \tilde{\V} \|^2_F  \text{,} \IEEEnonumber
	\end{IEEEeqnarray} 
	\begin{IEEEeqnarray}{rlll}
		\V^{(1,2)} = \text{arg} \underset{\tilde{\V} \in \mathcal{C}_2}{\min} \| \V^{(1)} - \tilde{\V} \|^2_F  \text{.} \IEEEnonumber
	\end{IEEEeqnarray} \IEEEnonumber
	The two projections are implemented element-wise as in Eq. (\ref{Pi_D}) and Eq. (\ref{Pi_N}), respectively.
	%
	% REFERENCES
	%
	\ifCLASSOPTIONcaptionsoff
	\newpage
	\fi
	\bibliographystyle{IEEEtran}
	\bibliography{references.bib}
	
\end{document}